\documentclass[sigconf]{acmart}

\AtBeginDocument{%
  }

\copyrightyear{2026}
\acmYear{2026}
\setcopyright{cc}
\setcctype{by}
\acmConference[WWW '26]{Proceedings of the ACM Web Conference 2026}{April 13--17, 2026}{Dubai, United Arab Emirates}
\acmBooktitle{Proceedings of the ACM Web Conference 2026 (WWW '26), April 13--17, 2026, Dubai, United Arab Emirates}
\acmPrice{}
\acmDOI{10.1145/3774904.3792639}
\acmISBN{979-8-4007-2307-0/2026/04}

\usepackage{amsmath,amssymb,amsfonts}
\usepackage{graphicx}
\usepackage{textcomp}
\usepackage{xcolor}
\usepackage{multicol}
\usepackage{amsthm}
\usepackage[linesnumbered,ruled]{algorithm2e}
\usepackage{caption}
\usepackage{subfigure}
\usepackage{url}
\usepackage{multirow}
\usepackage{float}
\usepackage{enumitem}
\usepackage{booktabs}

\usepackage[normalem]{ulem}
\usepackage{xcolor}
\usepackage{caption}

\usepackage{hyperref}
\usepackage{url}

\hypersetup{
    colorlinks=true,
    linkcolor=blue,
    citecolor=blue,
    urlcolor=blue
}
\begin{document}

\title{SEMixer: Semantics Enhanced MLP-Mixer for
Multiscale Mixing and Long-term Time Series
Forecasting}


\settopmatter{authorsperrow=4}
\author{Xu Zhang}
\orcid{0009-0006-5317-2422}
% \authornote{This work was conducted when Xu Zhang was interning at Ant Group. This work was supported by Ant Group Research Intern Program.}
\affiliation{%
  \institution{Shanghai Key Laboratory of Data Science, College of Computer Science and Artificial Intelligence\\ Fudan University}
  \city{Shanghai}
  \country{China}
}
\email{xuzhang22@m.fudan.edu.cn}

\author{Qitong Wang}
\authornote{Work done while at Université Paris Cité.}
\authornote{Both are corresponding authors.}
% 9
\affiliation{%
  \institution{Harvard University}
  \city{Cambridge,Massachusetts}
  \country{United States}
}
\email{qitong@seas.harvard.edu}

\author{Peng Wang}
% \authornote{Peng Wang is the corresponding author.}
\authornotemark[2]
\affiliation{%
  \institution{Shanghai Key Laboratory of Data Science, College of Computer Science and Artificial Intelligence\\ Fudan University}
  \city{Shanghai}
  \country{China}
}
\email{pengwang5@fudan.edu.cn}

\author{Wei Wang}
\affiliation{%
  \institution{Shanghai Key Laboratory of Data Science, College of Computer Science and Artificial Intelligence\\ Fudan University}
  \city{Shanghai}
  \country{China}
  % \streetaddress{P.O. Box 1212}
  % \country{USA}
}
\email{weiwang1@fudan.edu.cn}

% \renewcommand{\shortauthors}{Xu Zhang et al.}

\renewcommand{\shortauthors}{Xu Zhang, Qitong Wang, Peng Wang, and Wei Wang}

\begin{abstract}

Modeling multiscale patterns is crucial for long-term time series forecasting (TSF). However, redundancy and noise in time series, together with semantic gaps between non-adjacent scales, make the efficient alignment and integration of multi-scale temporal dependencies challenging. To address this, we propose SEMixer, a lightweight multiscale model designed for long-term TSF. SEMixer features two key components: a Random Attention Mechanism (RAM) and a Multiscale Progressive Mixing Chain (MPMC). RAM captures diverse time-patch interactions during training and aggregates them via dropout ensemble at inference, enhancing patch-level semantics and enabling MLP-Mixer to better model multi-scale dependencies. MPMC further stacks RAM and MLP-Mixer in a memory-efficient manner, achieving more effective temporal mixing. 
It addresses semantic gaps across scales and facilitates better multiscale modeling and forecasting performance. 
We not only validate the effectiveness of SEMixer on 10 public datasets, but also on the \textit{2025 CCF AlOps Challenge} based on 21GB real wireless network data, where SEMixer achieves third place. The code is available at \url{https://github.com/Meteor-Stars/SEMixer}.

\end{abstract}




\maketitle

\section{Introduction}

\begin{figure*}[bt]
% \vspace{-0.25cm}
\centerline{\includegraphics[width=1.0\linewidth]{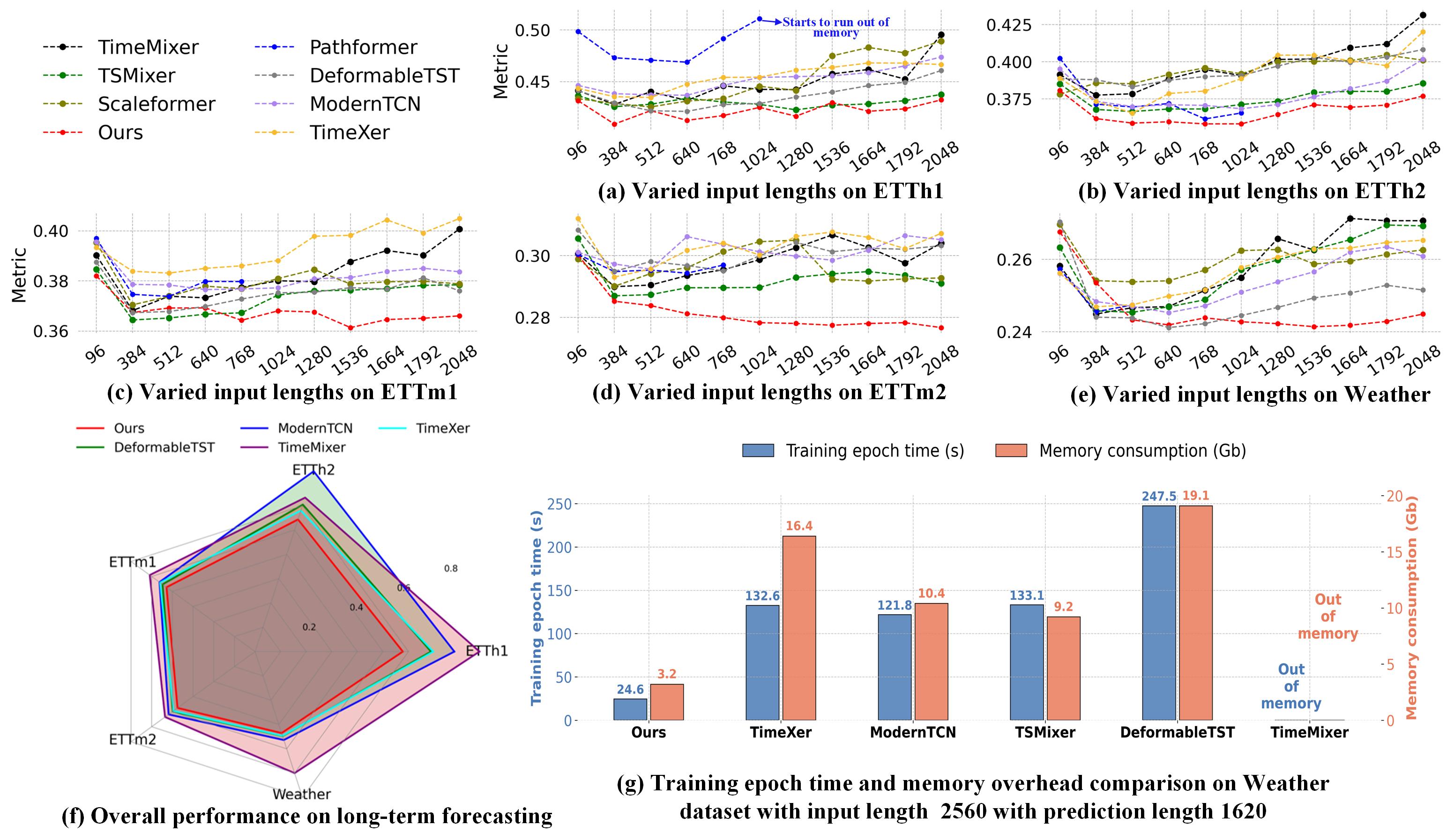}}
\vspace{-0.3cm} %调整的是Figure字样和图片的距离
\caption{\textbf{Sub-figures (a)-(e):} evaluate whether popular multi- and single-scale models benefit from longer historical sequences, reporting average MSE and MAE across 96, 192, 336, and 720 steps. \textbf{(f):} overall comparison on long-term forecasting. \textbf{(g):} training efficiency and memory overhead. SEMixer uses the same hyperparameters for all datasets and prediction lengths, adjusting only input length, demonstrating strong generalization and performance gains from longer sequences.}
\label{fig:moti_intro}
\vspace{-0.3cm}
\end{figure*}

Time series (TS) are ubiquitous on the World Wide Web, such as financial trading sequences~\cite{zhang2024self,xu2021rest} and performance monitoring metrics in microservice systems~\cite{jiang2023look}. Long-term time series forecasting (LTSF) plays a critical role in modern web technologies by leveraging historical data to predict future trends~\cite{zhang2025lightweight,hou2022multi,zhang2025multi,liu2024unitime}, enabling applications like web economics modeling~\cite{xu2021rest} and microservice log analysis~\cite{jiang2023look}. TS often exhibit diverse fluctuations across temporal scales~\cite{chen2024pathformer}, motivating multi-scale modeling for LTSF~\cite{chen2024pathformer,shabani2022scaleformer,wangtimemixer} to capture both short-term variations and long-term trends. For example, daily and monthly inputs capture long-term trends in electricity consumption, such as holiday effects and seasonality, which are difficult to identify at the hourly scale. Modeling interactions between daily/monthly and hourly patterns enables richer contextual and trend information for more accurate forecasts. However, despite their effectiveness, existing multi-scale models still face fundamental challenges that limit forecasting performance.

\textbf{First}, modeling multiple resolutions simultaneously introduces substantial computational overhead that grows rapidly with input length and model depth. This limits scalability and hinders the use of long historical sequences that are crucial for accurate real-world forecasting.
On a single 3090 GPU (24GB memory), the Transformer-based multiscale model Pathformer~\cite{chen2024pathformer} runs out of memory even on the small-scale ETTh dataset (Figure~\ref{fig:moti_intro}(a)). Although the MLP-based TimeMixer is more efficient, it still incurs significantly higher memory overhead than single-scale models such as TimeXer~\cite{wangtimexer}, ModernTCN~\cite{luo2024moderntcn}, and TSMixer~\cite{vijay2023tsmixer} (Figure~\ref{fig:moti_intro}(g)).
\textbf{Second}, time series often contain high redundancy and noise~\cite{pandit2017noise,nunes2023challenges,diebold2013correlation}, and non-adjacent scales (e.g., hourly vs. monthly) exhibit semantic gaps that reflect different underlying dynamics. As a result, effectively aligning and integrating multi-scale dependencies remains challenging. As shown in Figure~\ref{fig:moti_intro}(a)–(e), while existing multiscale models perform promisingly with short inputs, their performance degrades with longer inputs, suggesting that key signals may be diluted as input length increases.

To tackle these challenges, we propose SEMixer, a novel multi-scale model for efficient and effective long-term TSF. SEMixer builds on the widely adopted MLP-Mixer~\cite{tolstikhin2021mlp} in the Computer Vision (CV) domain, which is lightweight and leverages MLPs with permutation operations to mix features within and across patches, capturing both short- and long-term dependencies without relying on self-attention. To fully unlock MLP-Mixer’s potential for long-term TSF, SEMixer introduces two key components: the Random Attention Mechanism (RAM), which enriches the semantics of each time patch, and the Multiscale Progressive Mixing Chain (MPMC), which effectively models multi-scale temporal dependencies while addressing semantic gaps across different scales.

One key to MLP-Mixer’s success in CV is that image patches have rich semantics. In time series, however, time patches are sparser (e.g., only rising or falling trends) and may contain noise, limiting representation after mixing. To address this, we enhance time patch semantics via interactions. For encoded $N$ time patches $\mathcal{X}_d \in \mathbb{R}^{N \times D}$ ($D$ is the embedding size), self-attention (SAM) can propagate information as $A\mathcal{X}_d$ with $A = \text{softmax}(QK^T/\sqrt{D})$, leveraging patch associations to uncover useful semantics, thereby supplementing and enhancing semantics in the original patch. However, SAM is costly for multi-scale inputs. To efficiently enhance semantics, we propose the Random Attention Mechanism (RAM).

Compared with SAM, (i) RAM replaces the attention score matrix with a simple 0–1 interaction matrix and avoids query–key computation, achieving much higher efficiency; (ii) RAM learns a large number of interaction patterns via random interactions during training and integrates them through dropout ensemble~\cite{srivastava2014dropout,DBLP:conf/nips/BaldiS13,DBLP:conf/icann/HaraSS16} during inference to enhance semantics.
This high interaction diversity enables RAM to capture sufficient and effective patterns, whereas SAM may overfit noise and degrade performance at both time-point~\cite{zeng2023transformers_linear} and patch levels~\cite{kim2024self}, partly because efficiency constraints and convergence difficulty limit it to adopt only a few attention heads.
We also observe that RAM facilitates adequate mixing between features of different scales, thereby further enhancing the performance of MLP-Mixer.

Finally, to address the semantic gap across different scales and better capture multi-scale temporal dependencies, we propose the Multiscale Progressive Mixing Chain (MPMC), which stacks RAM and MLP-Mixer to progressively and pairwise concatenate adjacent scale inputs for multi-scale mixing, starting from the finest scale and gradually transitioning to the coarsest scale.
Compared with jointly processing all scales, this progressive strategy encourages effective interaction between neighboring scales, improves robustness to noise, reduces memory overhead, and enables the use of longer input sequences. As shown in Figure~\ref{fig:moti_intro}, extensive experiments demonstrate that SEMixer benefits more from longer sequences than existing models (Figure~\ref{fig:moti_intro}(a)–(f)) and achieves higher efficiency with lower memory cost than both single-scale and multi-scale baselines (Figure~\ref{fig:moti_intro}(g)). Our contributions are as follows:

% 
% \begin{itemize}
\begin{enumerate}[noitemsep, topsep=0pt, wide=\parindent]
    \item We propose SEMixer, a lightweight multiscale model for long-term forecasting. With a carefully designed architecture, SEMixer can handle longer input sequences and more effectively leverage them to achieve superior performance.

    \item We propose the Random Attention Mechanism (RAM), which learns diverse interactions via random sampling during training and integrates them through dropout ensemble to enhance time-patch semantics, achieving higher efficiency and forecasting accuracy. 

    \item We propose the Multiscale Progressive Mixing Chain (MPMC), which progressively stacks RAM and the MLP-Mixer backbone across increasing time-series scales and applies them only to pairwise concatenations of adjacent scales. This design considers semantic gaps between scales, improves forecasting performance, reduces memory usage by avoiding joint processing of all scales, and empirically enhances noise robustness.

    \item We conduct extensive experiments on 10 popular benchmark datasets, a large-scale competition dataset, and 12 advanced baselines, showing that SEMixer outperforms both multiscale and single-scale TSF models in accuracy and efficiency. 
    Additionally, SEMixer achieves third place in the \textit{2025 CCF AlOps Challenge}.
\end{enumerate}
% \end{itemize}

\section{Related work}

\subsection{Single scale for long-term forecasting}
Thanks to the ability of the self-attention mechanism (SAM) to model long-term dependencies, a series of long-term forecasting models based on the Transformer are proposed, e.g., Pathformer~\cite{chen2024pathformer}, MLF~\cite{zhang2025multi}, and PatchTST~\cite{nie2022time_patchformer}. 
However, some studies have pointed out that the high-cost SAM may not be necessary in TSF~\cite{zeng2023transformers_linear,kim2024self}. 
A series of highly efficient and competitive models have been proposed, including simple linear layers or MLPs based models~\cite{zhang2025lightweight,chen2023tsmixer}, convolutional neural network (CNN) based models~\cite{luo2024deformabletst,luo2024moderntcn}, and cross-attention focused models~\cite{zhang2023crossformer,kim2024self}.

\subsection{Multiscale for long-term forecasting}
Most works focus on single-scale models, while multi-scale forecasting models are less common, and can be divided into Transformer-based and MLP-based approaches.

\textbf{Transformer-based multiscale models.} Pyraformer~\cite{liu2021pyraformer} uses downsampling to build multiscale representations, while Scaleformer~\cite{shabani2022scaleformer} further refines the structure by progressively forecasting at increasingly finer scales. PathFormer~\cite{chen2024pathformer} adaptively extracts and aggregates multiscale features based on temporal dynamics.

\textbf{MLP-based multiscale models.} FiLM~\cite{zhou2022film} integrates multiscale information via Legendre polynomials and Fourier projection. TimeMixer~\cite{wangtimemixer} decomposes series into trend and seasonal components for mixing, while TimeMixer++~\cite{wang2024timemixer++} transforms multiscale series into multi-resolution time images and uses 2D-CNN with dual-axis SAM, with inevitably high memory and computational overhead on long sequences.

\textbf{Notably}, while multi-scale approaches exist, we are the first to propose the MPMC architecture, which addresses semantic gaps and model overhead across scales, offering significant accuracy and efficiency advantages for long-term forecasting with longer inputs.

\section{SEMixer}
\begin{figure*}[h]
% \vspace{-0.2cm}
% \setlength{\abovecaptionskip}{0.1cm} Analysis of single step forecasting on Fund dataset.
\centerline{\includegraphics[width=1\linewidth]{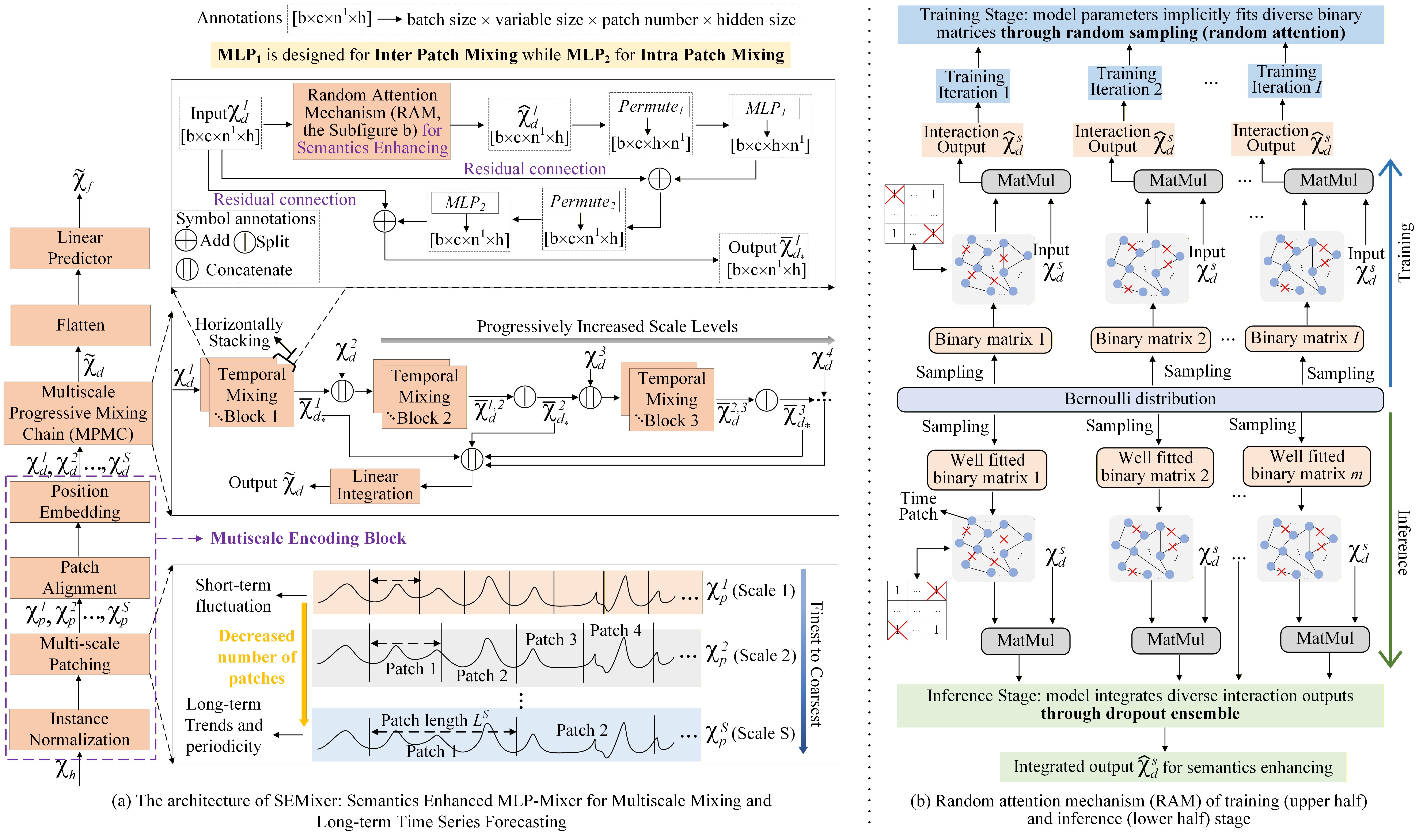} }
\vspace{-0.2cm}
\caption{SEMixer components: The Multiscale Encoding Block processes historical input $\mathcal{X}_h$ into $S$ multiscale inputs $\mathcal{X}_d^1, \mathcal{X}_d^2, ..., \mathcal{X}_d^S$. The MPMC structure then progressively performs temporal mixing on the finest-to-coarsest multiscale inputs to capture the multiscale temporal dependence. 
Each Temporal Mixing Block includes RAM (sub-figure b), Inter Patch Mixing ($\textit{Permute}_1$+$\textit{MLP}_1$), and Intra Patch Mixing ($\textit{Permute}_2$+$\textit{MLP}_2$). The multiscale outputs from all blocks are integrated for future forecasting.
}
\vspace{-0.3cm}
\label{fig:framework_MShyperv3}
\end{figure*}

We illustrate the proposed SEMixer in Figure~\ref{fig:framework_MShyperv3}. 
Next, we will detail each component of SEMixer.

\subsection{Multiscale encoding block}

\textbf{Instance normalization.} We first apply Instance Norm~\cite{DBLP:conf/iclr/KimKTPCC22} for historical input to address the distribution shift between training and testing data, producing $\mathcal{X}_h = [x_1, x_2, \ldots, x_n] \in \mathbb{R}^{n \times c}$ with length $n$ and $c$ variables.

\textbf{Multi-scale patching.} 
Then, given the $\mathcal{X}_h$, we generate multiscale inputs at the patch level to preliminarily enhance the semantics of time series.
Specifically, the $\mathcal{X}_h$ is progressively patchified across $S$ varied patch lengths $L$ and strides $K$, producing the multiscale inputs $\{\mathcal{X}_p^1, \dots, \mathcal{X}_p^s, \dots, \mathcal{X}_p^S\}$:
 \begin{equation}
 % \vspace{-0.1cm}
\label{equ:patch_cons}
\mathcal{X}_p^s=\text{Patchify}(\mathcal{X}_h,L^s,K^s),\ \ \mathcal{X}_p^s \in \mathbb{R}^{N^s \times L^s \times c},\ \ s\in\{1,...,S\}
% \vspace{-0.05cm}
\end{equation}
 \begin{equation}
 % \vspace{-0.1cm}
\label{equ:patch_cons}
L^s=\alpha^s \times L^1, K^s=\frac{L^s}{2},\ \ s\in\{2,...,S\}
% \vspace{-0.05cm}
\end{equation}
 \begin{equation}
 % \vspace{-0.1cm}
\label{equ:patch_cons}
N^s=\lfloor \frac{(n-L^s)}{K^s} \rfloor +2,\ \ s\in\{1,...,S\}
% \vspace{-0.05cm}
\end{equation}
where $\alpha^s$ is the scale factor for the $s$-th scale. The finest scale input has patch length $L^1$ and stride $K^1=L^1/2$. Once $L^1$ and $K^1$ are set, the patch length $L^s$ and stride $K^s$ for other scales are computed using $\alpha^s$. Here, $N^s$ denotes the number of scale patches and $\lfloor \cdot \rfloor$ denotes rounding down.

The patch length $L^s$ sets each patch’s sequence length, while the stride $K^s$ controls overlap between consecutive patches. If $K^s=L^s$, patches do not overlap, as shown in bottom of Figure~\ref{fig:framework_MShyperv3}(a). Different $L^s$ and $K^s$ generate multiple patched scales, providing varied temporal resolutions of $\mathcal{X}_h$. Finer scales (e.g., $\mathcal{X}_p^1$) with smaller $L$ and $K$ capture local patterns and high-frequency fluctuations, while coarser scales (e.g., $\mathcal{X}_p^S$) capture global trends and periodicity.

\textbf{Patch alignment and position embedding.}
Next, each scale input $\mathcal{X}_p^s$ is first linearly projected from patch length $L^s$ to dimension $D$ through the Patch Alignment module, enabling pairwise concatenation for the subsequent Multiscale Progressive Mixing Chain to extract multiscale temporal features.
Since the random attention mechanism ignores positional information, we add learnable position embeddings to each $\mathcal{X}_p^s$ after patch alignment to better enhance semantics:

\begin{equation}
\label{equ:patch_align_pos}
\mathcal{X}_d^{s}= \mathcal{X}_p^{s} W_{p}^{s}+W_{pos}^{s},\ \ s\in\{1,\dots,S\},
\end{equation}
where $W_p^s \in \mathbb {R}^{L^s \times D}$, $W^s_{pos} \in \mathbb {R}^{N^{s} \times D}$, and $\mathcal{X}_d^{s} \in \mathbb {R}^{N^{s} \times D}$.

\subsection{Random attention mechanism (RAM) to enhance semantics in time patch}

For clarity, we use the $s$-th scale $\mathcal{X}_d^s$ to illustrate RAM. In practice, RAM operates on pairwise concatenated adjacent-scale inputs.

\textbf{Training stage of RAM.}  
We first define a full interaction matrix $\widehat{A}^s \in \mathbb{R}^{N^s \times N^s}$ with all elements set to 1, indicating interactions between any two patches. 
During each training iteration, RAM randomly samples a binary matrix $M \in \{0,1\}^{N^{s}\times N^{s}}$ from a Bernoulli distribution as a mask to randomly cut off the interactions between patches. $M \sim \mathcal{B}(p)$, where $p$ (fixed at 0.85 in this paper) denotes the probability with which a connection is cut off between the $i$-th and $j$-th patches (e.g., $M_{ij}=0$ denotes red forks in Figure~\ref{fig:framework_MShyperv3}b). 
RAM forces the model’s learnable parameters to implicitly fit randomly sampled binary matrices, thereby inducing diverse interactions among patches.
After training convergence, these well-fitted binary matrices effectively enhance patch semantics and improve the performance of MLP-Mixer.
Over a total of $I$ training iterations, RAM can be formulated as follows:
\begin{equation}
\mathcal{\widehat{X}}_d^{s}= (M^1 \odot \widehat{A}^s )\mathcal{X}_d^{s}, \ \ 
\mathcal{\widehat{X}}_d^{s}= (M^2\odot \widehat{A}^s)\mathcal{X}_d^{s}, \ \ \dots, \ \ 
\mathcal{\widehat{X}}_d^{s}= (M^I\odot \widehat{A}^s)\mathcal{X}_d^{s}.
\label{equ:hyper_training}
\end{equation}
where $\mathcal{\widehat{X}}_d^{s}$ is the semantically enhanced $\mathcal{X}_d^{s}$.

\textbf{Inference stage of RAM.}
During inference, interaction outputs are integrated by sampling and aggregating all binary interaction matrices learned during training:
\begin{equation}
\mathcal{\widehat{X}}_d^{s}= \frac{1}{m} \sum_{i=1}^{m} M^i \widehat{A}^s \mathcal{X}_d^{s}.
\label{equ:hyper_inference_or}
\end{equation}

\textbf{Dropout ensemble.} Fortunately, RAM follows a similar principle to Dropout~\cite{srivastava2014dropout}. 
Dropout randomly deactivates neurons during training, implicitly training multiple subnetworks.
During inference, all neurons are active, but the output is scaled by the keep probability to maintain consistent output expectations between training and inference stages, which approximates the ensemble effect of multiple subnetworks~\cite{DBLP:conf/nips/BaldiS13,DBLP:conf/icann/HaraSS16}. 
Here, since $p$ denotes the disconnection probability, the keep probability is $1-p$. Following the spirit of Dropout, Eq.~\ref{equ:hyper_inference_or} can be approximated by:
\begin{equation}
\mathcal{\widehat{X}}_d^{s}= \widehat{A}^s \mathcal{X}_d^{s}\cdot(1-p).
\label{equ:hyper_inference}
\end{equation}
The training and inference stages of RAM are illustrated in Figure~\ref{fig:framework_MShyperv3}(b).
Compared with SAM, RAM is more efficient since it avoids query–key attention computation.
Unlike multi-head SAM with a limited number of heads, RAM fits and integrates a large number of interaction matrices (each can be viewed as a head), enabling it to learn sufficient effective interactions and achieve robust semantic enhancement without being dominated by noisy interactions. 

The enhanced $\mathcal{\widehat{X}}_d^{s}$ is then fed into inter- and intra-patch MLP-Mixers for feature mixing, thereby extracting multiscale temporal dependence (top of Figure~\ref{fig:framework_MShyperv3}(a)).
Together, RAM and MLP-Mixer form a temporal mixing block (middle of Figure~\ref{fig:framework_MShyperv3}(a)).

\textbf{Limitation and residual connection designs.}
RAM implicitly assumes meaningful correlations across patches and channels. Although it effectively enhances patch semantics on most datasets, it may introduce noise when such correlations are weak. To address this issue, we incorporate a residual design (top of Figure~\ref{fig:framework_MShyperv3}) that exploits beneficial correlations while preventing performance degradation when they are weak.

\begin{algorithm}[h]
\footnotesize 
\caption{Multiscale Progressive Mixing Chain (MPMC)}
    \label{alg:algorithm1}
    
    \KwIn{
    % \begin{enumerate}
  \textbf{(i)} $S$ \textit{TemporalMixing} blocks and $S$ multiscale embeddings $\mathcal{X}_d^1$,..., $\mathcal{X}_d^S$. \textbf{(ii)} Functions \textit{Concat}$(\cdot)$ and \textit{Split}$(\cdot)$, where the latter separates the $s$-th scale feature (which is mixed with the $(s-1)$-th scale) through slicing operations.
  % \end{enumerate}
 }
        % \KwOut{Multiscale features $O_{mixing}=\{\mathcal{\bar{X}}_{d_*}^1$,..., $\mathcal{\bar{X}}_{d_*}^S$\} after temporal mixing.}

        \KwOut{Multiscale features $O_{mixing}=\{\mathcal{\bar{X}}_{d_*}^1,\ldots,\mathcal{\bar{X}}_{d_*}^S\}$ after temporal mixing.}
    \BlankLine
        $O_{mixing} \Leftarrow \emptyset$
        
        \For{$s=1$ to $S$}{
            \eIf{s=1}
            {
            $\mathcal{\bar{X}}_{d_*}^1$=\textit{TemporalMixing}($\mathcal{X}_d^1$)

            $O_{mixing} \Leftarrow O_{mixing} \cup \mathcal{\bar{X}}_{d_*}^1$ 

            }
            {
            
            $\mathcal{\bar{X}}_{d}^{s-1,s}$=\textit{TemporalMixing}$\left( \textit{Concat}(\mathcal{\bar{X}}_{d_*}^{s-1},\mathcal{X}_d^s)\right)$

            $\mathcal{\bar{X}}_{d_*}^s$=\textit{Split}($\mathcal{\bar{X}}_{d}^{s-1,s}$)

          $O_{mixing} \Leftarrow O_{mixing} \cup \mathcal{\bar{X}}_{d_*}^s$ 
            }}

            \Return $O_{mixing}$
            % \Return {$\frac{\mathcal{L}}{K}$}
\end{algorithm}

\subsection{Multiscale progressive mixing chain (MPMC)}
This section proposes MPMC to stack the RAM and MLP-Mixers, as illustrated in the middle of Figure~\ref{fig:framework_MShyperv3}(a).
Specifically, MPMC performs temporal mixing through pairwise concatenation of adjacent scale inputs and progressively increases the number of involved scales to establish interactions across different scales.
Compared with directly concatenating all scale inputs, this pairwise strategy reduces memory overhead for long-sequence processing and mitigates semantic noise caused by pattern discrepancies between non-adjacent scales, enabling the model to better learn and distinguish multi-scale representations and capture temporal dependencies.

MPMC is formulated in Algorithm~\ref{alg:algorithm1} and produces multiscale features $O_{mixing}$, which are then concatenated and fed into a linear layer for dimensionality reduction and feature integration, yielding $\widetilde{X}_d$.
Finally, $\widetilde{X}_d$ is flattened and passed to a linear predictor to forecast future values $\mathcal{X}_f=[x_{n+1},x_{n+2},...,x_{n+t}] \in \mathbb {R}^{t \times C}$ for all $c$ variables.
Following most existing works, we adopt the MSE loss function for model optimization.

\section{Experiments}
\label{sec:exp_set}
\subsection{Experimental settings}
\subsubsection{\textbf{Datasets and evaluation metrics.}} We validate our method on 10 widely used public long-term forecasting datasets~\cite{nie2022time_patchformer,wangtimemixer,chen2024pathformer} across industry (4 ETT datasets), climate (Weather), energy (Solar Energy, Electricity), health (Influenza-Like Illness, ILI), economy (Exchange), and traffic. Following prior works, we use MSE and MAE for multivariate long-term TSF evaluation. \textbf{We also demonstrate the effectiveness of SEMixer on the 2025 CCF AlOps Challenge.} The training set contains ~26,000 4G wireless cells (21GB) over three weeks at 15-minute intervals. The task is to forecast next-day Key Performance Indicator (KPI) trends (96 steps) for 1,000 cells to aid network maintenance, thereby improving service quality and user experience. More details are available at\footnote{\url{https://challenge.aiops.cn/home/competition/1920775077574500373}}. 

\subsubsection{\textbf{Diverse baselines.}}
% \textbf{Baselines.}
We compare SEMixer with 12 advanced baselines across five categories:
\textbf{(i) Multi-scale transformers}: Pathformer~\cite{chen2024pathformer}, Scaleformer~\cite{shabani2022scaleformer};
\textbf{(ii) Multi-scale linear models}: FiLM~\cite{zhou2022film}, TimeMixer~\cite{wangtimemixer};
\textbf{(iii) Single-scale transformers}: TimeXer~\cite{wangtimexer}, PatchTST~\cite{nie2022time_patchformer}, iTransformer~\cite{DBLP:conf/iclr/LiuHZWWML24};
\textbf{(iv) Single-scale linear models}: TSMixer~\cite{vijay2023tsmixer}, DLinear~\cite{zeng2023transformers_linear};
\textbf{(v) CNN-based models}: DeformableTST~\cite{luo2024deformabletst}, ModernTCN~\cite{luo2024moderntcn}, TimesNet~\cite{wutimesnet}.
Comparisons with FiLM and Scaleformer are reported in Appendix Table~\ref{tab:metric_public_long_term_appendix_p2}.
\begin{table*}[h]
 % \vspace{-0.1cm}
    % \renewcommand{\arraystretch}{0.5}
    \setlength{\tabcolsep}{2.95pt}
    % {|>{\setlength{\tabcolsep}{3pt}}c|c|c|}
    \centering
    \caption{Forecasting results on public datasets and advanced baselines, averaged over 96, 192, 336, and 720 steps. Full results are provided in Appendix Table~\ref{tab:metric_public_long_term_appendix_p2}, including FiLM and Scaleformer. Best results are in bold, and second-best are underlined.}
    \vspace{-0.1cm}
    \label{tab:metric_public_long_term}
    % \begin{tabular}{c|c|p{20pt}p{20pt}|cc|cc|cc|cc|cc}
    {\footnotesize  
    \begin{tabular}{c|c|c|c|c|c|c|c|c|c|c|c|c|c|c|c|c|c|c|c|c|c|c}
        \hline
        \multirow{2}{*}{\shortstack{}}  &  \multicolumn{2}{c|}{\shortstack{SEMixer}} &  \multicolumn{2}{c|}{\shortstack{Deform.TST}} & \multicolumn{2}{c|}{\shortstack{TimeXer}} & \multicolumn{2}{c|}{\shortstack{ModernTCN}}& \multicolumn{2}{c|}{\shortstack{Pathformer}}
        &  \multicolumn{2}{c|}{\shortstack{iTransformer}}
        & \multicolumn{2}{c|}{\shortstack{TimesNet}}& \multicolumn{2}{c|}{\shortstack{TSMixer }}& \multicolumn{2}{c|}{\shortstack{DLinear}}& \multicolumn{2}{c|}{PatchTST}& \multicolumn{2}{c}{TimeMixer}\\
         &  MSE & MAE & MSE & MAE & MSE & MAE  & MSE & MAE & MSE & MAE& MSE & MAE& MSE & MAE& MSE & MAE& MSE & MAE& MSE & MAE& MSE & MAE\\ 
         \hline

\multirow{1}{*}{\rotatebox[origin=c]{0}{ETTh1}}&\textbf{0.400} &\textbf{0.418}&\underline{0.408} &\underline{0.429}&0.425 &0.444&0.426 &0.445&0.437 &0.441&0.439 &0.455&0.456 &0.448&0.412 &0.432&0.421 &0.442&0.413 &0.434&0.419 &0.435\\
\hline
\multirow{1}{*}{\rotatebox[origin=c]{0}{ETTh2}}&\textbf{0.331} &\textbf{0.382}&0.363 &0.402&0.345 &0.391&\underline{0.336} &\underline{0.39}&0.343 &0.392&0.366 &0.404&0.401 &0.426&0.341 &0.391&0.431 &0.447&0.34 &0.388&0.353 &0.395\\
\hline
\multirow{1}{*}{\rotatebox[origin=c]{0}{ETTm1}}&\textbf{0.342} &\textbf{0.375}&0.351 &0.383&0.366 &0.395&0.36 &0.39&0.365 &0.39&0.362 &0.396&0.392 &0.404&\underline{0.348} &\underline{0.378}&0.35 &0.38&0.347 &0.384&0.352 &0.383\\
\hline
\multirow{1}{*}{\rotatebox[origin=c]{0}{ETTm2}}&\textbf{0.241} &\textbf{0.312}&0.263 &0.322&0.263 &0.325&0.258 &0.324&0.258 &0.32&0.262 &0.33&0.289 &0.334&0.25 &0.316&0.255 &0.327&\underline{0.246} &\underline{0.315}&0.256 &0.318\\
\hline
\multirow{1}{*}{\rotatebox[origin=c]{0}{Weather}}&\textbf{0.216} &\textbf{0.258}&\underline{0.221} &\underline{0.26}&0.226 &0.266&0.224 &0.266&0.231 &0.269&0.236 &0.273&0.256 &0.286&0.224 &0.262&0.232 &0.281&0.223 &0.263&0.223 &0.264\\
\hline
\multirow{1}{*}{\rotatebox[origin=c]{0}{Electricity}}&\textbf{0.154} &\textbf{0.249}&0.161 &0.261&0.164 &0.269&\underline{0.156} &\underline{0.252}&0.17 &0.268&0.165 &0.266&0.192 &0.295&0.161 &0.255&0.162 &0.262&0.164 &0.258&0.164 &0.258\\
\hline
\multirow{1}{*}{\rotatebox[origin=c]{0}{ILl}}&\textbf{2.385} &\textbf{1.080}&2.799 &1.174&2.758 &1.161&2.898 &1.2&2.752 &1.145&\underline{2.585} &\underline{1.111}&3.597 &1.258&2.799 &1.174&2.758 &1.161&2.898 &1.2&2.752 &1.145\\
\hline
\multirow{1}{*}{\rotatebox[origin=c]{0}{Exchange}}&\textbf{0.344} &\textbf{0.397}&0.423 &0.434&0.386 &0.419&0.491 &0.464&0.55 &0.48&0.37 &0.426&0.602 &0.556&0.431 &0.438&\underline{0.373} &\underline{0.416}&0.385 &0.419&0.438 &0.45\\
\hline
\multirow{1}{*}{\rotatebox[origin=c]{0}{Solar Energy}}&\underline{0.184} &\underline{0.245}&\textbf{0.184} &\textbf{0.241}&0.19 &0.268&0.22 &0.306&0.24 &0.283&0.232 &0.3&0.231 &0.3&0.187 &0.246&0.23 &0.294&0.185 &0.247&0.214 &0.275\\
\hline
\multirow{1}{*}{\rotatebox[origin=c]{0}{Traffic}}&\textbf{0.388} &\textbf{0.268}&\underline{0.393} &\underline{0.277}&0.397 &0.284&0.394 &0.277&0.412 &0.296&0.416 &0.314&0.572 &0.308&0.393 &0.277&0.397 &0.284&0.394 &0.277&0.412 &0.296\\
\hline
    \end{tabular}
    }
\vspace{-0.1cm}
\end{table*}

\begin{table*}[h]
 % \vspace{-0.2cm}
    % \renewcommand{\arraystretch}{0.5}
    \setlength{\tabcolsep}{3.2pt}
    % {|>{\setlength{\tabcolsep}{3pt}}c|c|c|}
    \centering
    \caption{Forecasting with an input length of 2560 for predicting 1020, 1320, and 1620 steps, with results averaged across all horizons. SEMixer also achieves superior performance with a fixed input length of 2048. “–” indicates out-of-memory.}
    \vspace{-0.1cm}
    \label{tab:metric_public_ultra_long_term}
    % \begin{tabular}{c|c|p{20pt}p{20pt}|cc|cc|cc|cc|cc}
    %tiny
    {\footnotesize  
    \begin{tabular}{c|c|c|c|c|c|c|c|c|c|c|c|c|c|c|c|c|c|c|c|c|c|c}
        \hline
        \multirow{2}{*}{\shortstack{}}  &  \multicolumn{2}{c|}{\shortstack{SEMixer}} &  \multicolumn{2}{c|}{\shortstack{Deform.TST}} & \multicolumn{2}{c|}{\shortstack{TimeXer}} & \multicolumn{2}{c|}{\shortstack{ModernTCN}}
        &  \multicolumn{2}{c|}{\shortstack{iTransformer}}
        & \multicolumn{2}{c|}{\shortstack{TimesNet}}& \multicolumn{2}{c|}{\shortstack{TSMixer }}& \multicolumn{2}{c|}{\shortstack{DLinear}}& \multicolumn{2}{c|}{PatchTST}& \multicolumn{2}{c|}{TimeMixer}& \multicolumn{2}{c}{Scaleformer}\\
         &  MSE & MAE & MSE & MAE & MSE & MAE  & MSE & MAE & MSE & MAE& MSE & MAE& MSE & MAE& MSE & MAE& MSE & MAE& MSE & MAE& MSE & MAE\\ 
         \hline
\multirow{1}{*}{\rotatebox[origin=c]{0}{ETTh1}}&\textbf{0.595} &\textbf{0.554}&0.728 &0.612&0.752 &0.648&0.877 &0.691&0.856 &0.707&1.226 &0.874&0.633 &0.572&0.745 &0.645&\underline{0.627} &\underline{0.571}&0.933 &0.685&0.909 &0.713\\
\hline
\multirow{1}{*}{\rotatebox[origin=c]{0}{ETTh2}}&{0.551} &\textbf{0.536}&0.616 &0.577&0.602 &0.56&0.847 &0.675&0.634 &0.584&0.787 &0.65&0.565 &0.547&1.379 &0.809&\textbf{0.547} &\underline{0.541}&0.615 &0.575&0.591 &0.571\\
\hline
\multirow{1}{*}{\rotatebox[origin=c]{0}{ETTm1}}&\textbf{0.425} &\textbf{0.430}&0.448 &0.45&0.454 &0.466&0.463 &0.466&0.489 &0.485&- &-&\underline{0.431} &\underline{0.442}&0.428 &0.449&0.447 &0.452&0.506 &0.49&0.443 &0.453\\
\hline
\multirow{1}{*}{\rotatebox[origin=c]{0}{ETTm2}}&\textbf{0.355} &\textbf{0.398}&0.379 &0.416&0.374 &0.413&0.399 &0.434&0.419 &0.437&- &-&0.373 &0.414&0.477 &0.477&\underline{0.37} &\underline{0.409}&0.423 &0.439&0.388 &0.428\\
\hline
\multirow{1}{*}{\rotatebox[origin=c]{0}{Weather}}&\textbf{0.323} &\textbf{0.349}&0.335 &0.359&0.334 &0.36&0.352 &0.379&0.353 &0.376&- &-&0.338 &0.365&0.329 &0.366&\underline{0.33} &\underline{0.357}&- &-&0.338 &0.371\\
\hline

    \end{tabular}}
\vspace{-0.1cm}
\end{table*}
\subsubsection{\textbf{Implementation details.}}
The hidden size $D$ for patch and position embeddings is set to 128, while the hidden size $\widehat{\mathcal{X}}_d$ for linear integration in the temporal mixing block is set to 64.
The patch number $N^1$ at the finest scale is fixed at 64, and the number of scales $S$ is set to 4.
The scale factors $\alpha^2$, $\alpha^3$, and $\alpha^4$ for computing $L^s$ and $K^s$ are fixed at 2, 4, and 8, respectively.
The sampling disconnection probability $p$ for binary matrices is set to 0.85. 

We observe that some models perform well with short inputs (e.g., 96) but degrade with longer input lengths. To ensure fair comparison and evaluate the baselines' ability to benefit from longer inputs, we not only evaluate with a fixed input length (512, 2048 and 2560) but also search for the optimal input length from \{96, 384, 512,640,
768, 1024, 1280, 1536, 1664, 1792, 2048\}, which is obtained by scaling forward or backward from the baseline length of 512 using predefined ratio factors.
All models are trained for 30 epochs (50 for Traffic), using identical data loading settings, and run on an NVIDIA GeForce RTX 3090 GPU with PyTorch.

\subsection{Main results}

Figure~\ref{fig:moti_intro} shows that SEMixer achieves higher accuracy from longer inputs than existing methods, with better efficiency and lower memory cost, owing to its two key designs: MPMC and RAM. We omit the advantages already illustrated in Figure~\ref{fig:moti_intro}. The time complexity analysis of RAM is provided in Appendix~\ref{sec:time_complexity}. Next, we present more results with multiple seeds to ensure reliable conclusions.

\subsubsection{\textbf{Long-term forecasting results}}
Table~\ref{tab:metric_public_long_term} shows that SEMixer achieves superior long-term forecasting performance, with 5–15\% lower MSE than existing models, demonstrating its ability to capture complex temporal dependencies across diverse time series.
The MPMC design efficiently handles long sequences by modeling both short-term fluctuations and long-term trend–periodicity, while the progressive interaction chain mitigates semantic gaps across scales.
In addition, RAM enhances time-patch semantics through multi-scale random interactions, further unlocking the potential of MLP-Mixer for TSF.
When extending input and prediction lengths (Table~\ref{tab:metric_public_ultra_long_term}), SEMixer consistently maintains significant performance advantages, confirming its robustness and effectiveness.

\subsubsection{\textbf{Performance comparisons on \textit{2025 CCF AlOps Challenge}}}
We apply SEMixer to the wireless network prediction task in the \textit{2025 CCF AlOps Challenge} and achieve third place.
During the competition, we further compared SEMixer with several advanced methods. As shown in Table~\ref{tab:ccf_aiops}, SEMixer attains the lowest prediction error among all competitors.
All results are obtained by submitting model predictions to the official evaluation platform.
To ensure a fair comparison, all experimental settings (e.g., batch size, learning rate, input length, and loss function) are kept identical, with only the model architectures varied.

\begin{table}[bt]
    \setlength{\tabcolsep}{2.5pt}
    % {|>{\setlength{\tabcolsep}{3pt}}c|c|c|}
    \centering
    \caption{Model prediction errors on the wireless network KPI forecasting dataset in \textit{2025 CCF AlOps Challenge}.  }
    \vspace{-0.2cm}
    \label{tab:ccf_aiops}
    
{\small
    \begin{tabular}{c|c|c|c|c|c|c}
        \hline
        \multirow{1}{*}{\shortstack{}}  &  \multicolumn{1}{c|}{SEMixer} &  \multicolumn{1}{c|}{Deform.TST}&  \multicolumn{1}{c|}{TimeXer}  & \multicolumn{1}{c|}{ModernTCN} & \multicolumn{1}{c|}{PatchTST}& \multicolumn{1}{c}{TimeMixer} \\
         % & & MSE  & MSE  & MSE  & MSE \\ 
         \hline
\multirow{1}{*}{\rotatebox[origin=c]{0}{}}&\textbf{0.4425}&0.4491&0.4486&0.4482&0.4506&\underline{0.4479}\\

\hline
    \end{tabular}}
    \vspace{-0.3cm}
\end{table}

\begin{table*}[bt]
    \setlength{\tabcolsep}{2.8pt}
    % {|>{\setlength{\tabcolsep}{3pt}}c|c|c|}
    \centering
    \caption{Ablation study, reporting the average MSE over horizons 96, 192, 336, and 720. SAM, PS, AC, FA, LS, LSH, and Performer denote different self-attention mechanisms: standard self-attention, ProbSparse~\cite{zhou2021informer}, AutoCorrelation~\cite{wu2021autoformer}, FourierAttention~\cite{zhou2022fedformer}, LogSparse~\cite{li2019enhancing}, Locality-Sensitive Hashing~\cite{kitaevreformer}, and Performer attention~\cite{choromanskirethinking}. The last two rows report per-epoch time and memory usage for comparing the efficiency of different attention mechanisms.  “–” indicates out-of-memory. }
    \vspace{-0.2cm}
    \label{tab:ablation_study_attention}
    
{\small
    \begin{tabular}{c|c|c|c|c|c|c|c|c|c|c}
        \hline
        \multirow{1}{*}{\shortstack{}}  &  \multicolumn{1}{c|}{SEMixer} &  \multicolumn{1}{c|}{w/o MPMC}&  \multicolumn{1}{c|}{w/o RAM}  & \multicolumn{1}{c|}{w/ SAM} & \multicolumn{1}{c|}{w/ PS}& \multicolumn{1}{c|}{w/ AC}& \multicolumn{1}{c|}{w/ FA}& \multicolumn{1}{c|}{w/ LS}& \multicolumn{1}{c|}{w/ LSH}& \multicolumn{1}{c}{w/ Performer}  \\
         % & & MSE  & MSE  & MSE  & MSE \\ 
         \hline
\multirow{1}{*}{\rotatebox[origin=c]{0}{ETTh1}}&\textbf{0.400}&\underline{0.405}&0.411&0.419&0.414&0.412&0.411&0.413&0.409&0.411\\
\hline
\multirow{1}{*}{\rotatebox[origin=c]{0}{ETTh2}}&\textbf{0.331}&0.34&0.336&0.336&0.334&0.334&0.334&\underline{0.333}&0.335&0.335\\
\hline
\multirow{1}{*}{\rotatebox[origin=c]{0}{ETTm1}}&\textbf{0.342}&0.347&0.347&0.36&\underline{0.346}&0.349&0.346&0.346&0.346&-\\
\hline
\multirow{1}{*}{\rotatebox[origin=c]{0}{ETTm2}}&\textbf{0.240}&\underline{0.246}&0.249&0.258&0.25&0.251&0.246&0.252&0.252&-\\
\hline
\multirow{1}{*}{\rotatebox[origin=c]{0}{Weather}}&\textbf{0.216}&0.223&0.227&0.225&\underline{0.220}&0.234&0.225&0.219&0.227&-\\
\hline
\multirow{1}{*}{\rotatebox[origin=c]{0}{Electricity}}&\textbf{0.154}&0.158&\underline{0.156}&-&-&-&0.16&-&-&-\\
\hline
\hline
\multirow{1}{*}{\rotatebox[origin=c]{0}{\shortstack{Weather}}}&\textbf{\shortstack{21.42s \\(2.96GB)} }&\shortstack{10.43s \\(2.06GB)}&\shortstack{20.26s \\(2.91GB)}&\shortstack{166.94s \\(17.74GB)}&\shortstack{64.43s \\(4.66GB)}&\shortstack{46s \\(5.49GB)}&\shortstack{100.77s \\(4GB)}&\shortstack{78.55s \\(8.38GB)}&\shortstack{336.44s \\(11.27GB)}&-\\
\hline
\multirow{1}{*}{\rotatebox[origin=c]{0}{Electricity}}&\textbf{\shortstack{128.43s \\(15.99GB)} }&\shortstack{56.43s \\(7.75GB)}&\shortstack{126.6s \\(15.79GB)}&-&\shortstack{-}&-&\shortstack{653.63s \\(22.16GB)} &-&-&-\\
\hline
    \end{tabular}}
    % \vspace{-0.1cm}
\end{table*}

\subsection{Ablation study}
\begin{figure*}[bt]
% \vspace{-0.25cm}
\centerline{\includegraphics[width=1.0\linewidth]{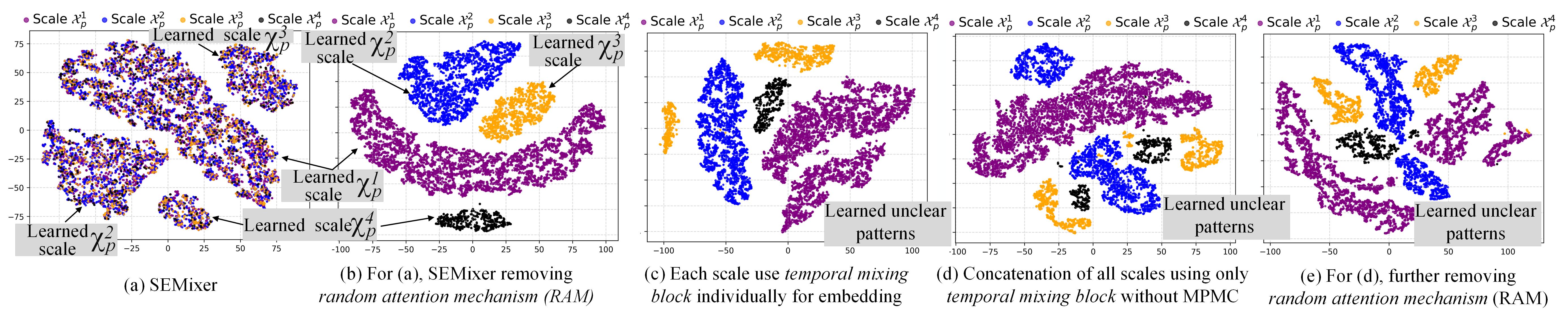}}
\vspace{-0.2cm} %调整的是Figure字样和图片的距离
\caption{Visualizing learned 12000 patch embeddings of 4 scale inputs $\mathcal{X}_p^1$,$\dots$, $\mathcal{X}_p^4$ on Weather test set with predicting length 720.}
\label{fig:vis_tsne}
\vspace{-0.1cm}
\end{figure*}

\begin{figure}[bt]
% \vspace{-0.25cm}
\centerline{\includegraphics[width=0.8\linewidth]{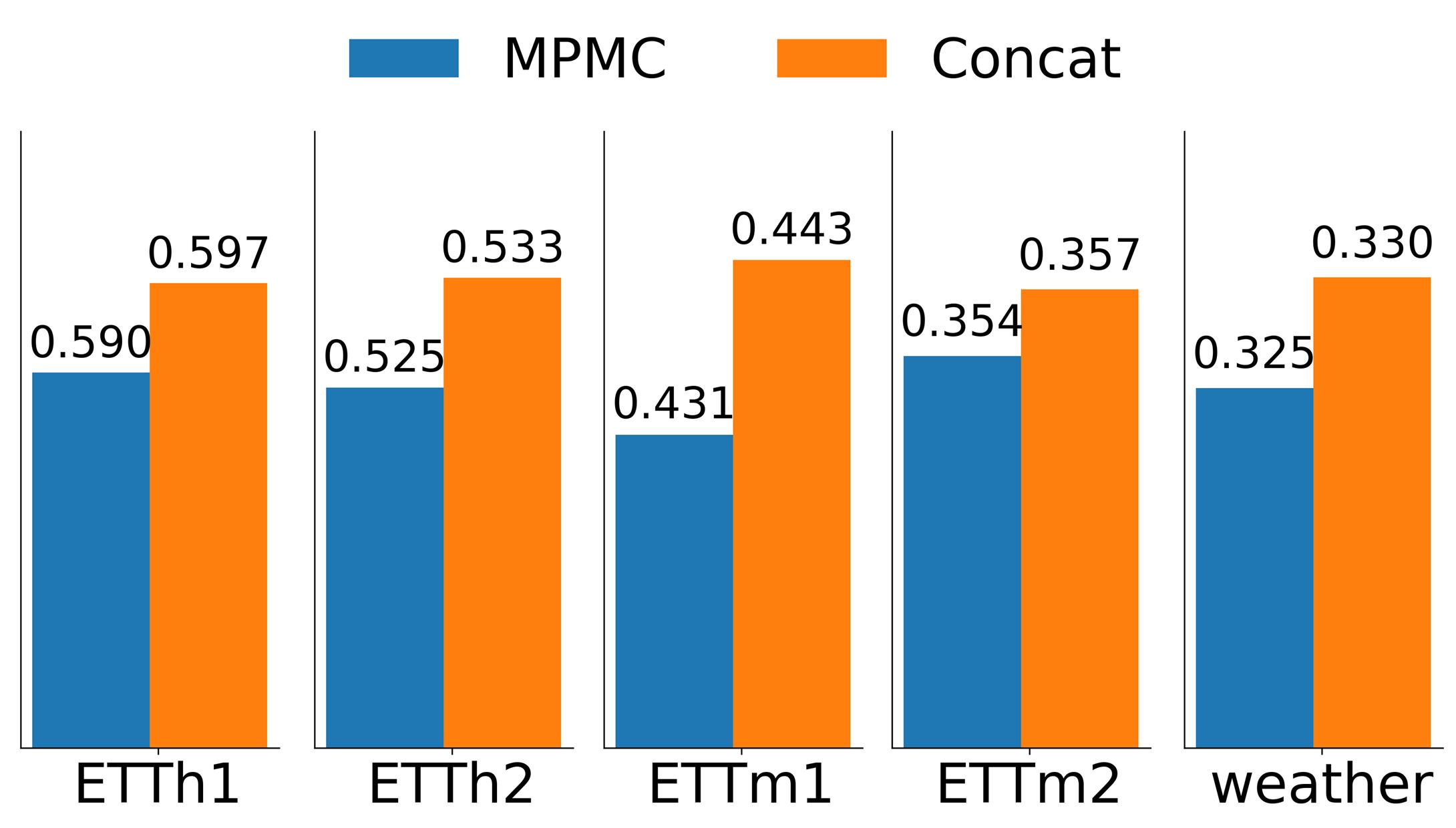}}
\vspace{-0.2cm} %调整的是Figure字样和图片的距离
\caption{Further ablation study of MPMC structure. Average MSE across all prediction lengths (1020, 1320 and 1620) under 2048 and 2560 sequence lengths on the long-term TSF task.}
\label{fig:mpmc_concat}
\vspace{-0.3cm}
\end{figure}

We evaluate the effectiveness of SEMixer through both quantitative and qualitative analyses.
\textbf{Quantitatively}, we remove or replace key components and examine the resulting performance changes, as shown in Table~\ref{tab:ablation_study_attention}, Table~\ref{tab:ultra_long_ablation_2560_overall}, and Figure~\ref{fig:mpmc_concat}.
\textbf{Qualitatively}, we illustrate the effectiveness of RAM and MPMC via patch embedding visualizations in Figure~\ref{fig:vis_tsne}.

\subsubsection{\textbf{Random attention mechanism}}

\textbf{(1)} As shown in Table~\ref{tab:ablation_study_attention}, removing the random attention mechanism (w/o RAM, i.e., removing Eq.~\ref{equ:hyper_training} and Eq.~\ref{equ:hyper_inference}) increases the MSE by about 5\%, indicating that RAM effectively enhances the performance of MLP-Mixer. By modeling associations among multiscale time patches, RAM uncovers richer semantics in time series data. For example, integrating local patch features (e.g., trends and periodicity) with information from other patches yields more complete global semantics, such as overall trend turning points and anomalies. Consequently, semantic enhancement across patches is crucial for capturing key temporal patterns in long sequences with increasing noise redundancy. Similar conclusions are also observed in the extended setting with longer input lengths (Table~\ref{tab:ultra_long_ablation_2560_overall}).

\textbf{(2)} To validate RAM, we replace it with various self-attention mechanisms and compare both predictive performance and training overhead. As shown in Table~\ref{tab:ablation_study_attention}, these replacements lead to clear performance degradation and higher overhead, consistent with prior studies~\cite{zeng2023transformers_linear,kim2024self}. This indicates that standard attention may capture noisy or redundant correlations due to limited heads. In contrast, RAM learns diverse interactions through random sampling and expectation-based integration, enabling more effective semantic extraction. The residual design further prevents performance degradation when noisy interactions arise. Moreover, RAM avoids QKV computations, achieving up to threefold higher efficiency and much lower memory usage, e.g., compared with ProbSparse attention (PS).

\textbf{(3)} We add a dropout layer after the standard SAM's attention scores with a dropout rate of 0.85. On ETTh1, ETTh2, ETTm1, ETTm2, and Weather, the average performance is
0.417±0.0025, 0.336±0.001, 0.352±0.0046, 0.254±0.001, and 0.221±0.0013, respectively.
Although this brings improvements, the performance remains inferior to ours (0.400±0.0005, 0.331±0.0007, 0.342±0.0011, 0.240±0.0016, 0.216±0.0025, 0.154).
Moreover, this approach still requires computing attention scores via Q–K–V, incurring substantial computational overhead.

\begin{figure*}[h!]
% \vspace{-0.25cm}
\centerline{\includegraphics[width=1.0\linewidth]{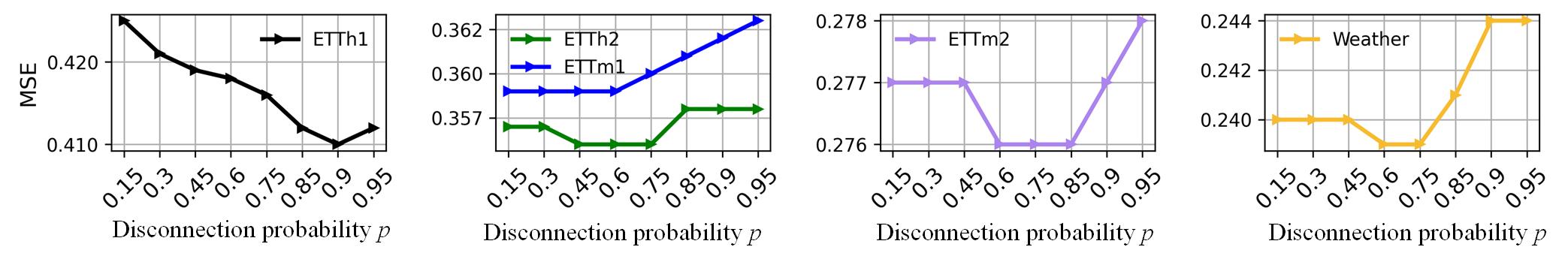}}
\vspace{-0.2cm} %调整的是Figure字样和图片的距离
\caption{Hyperparameter analysis of the random disconnection probability $p$ on various datasets.}
\label{fig:Hyperparameter}
\vspace{-0.1cm}
\end{figure*}

\subsubsection{\textbf{Multiscale progressive mixing chain}}

Thanks to MPMC's progressive interaction, MLP-Mixer does not need to process all scale inputs at once, which can obviously reduce memory overhead. Next, we further verify the impact of MPMC on prediction accuracy.

\textbf{(1)} As shown in Table~\ref{tab:ablation_study_attention}, removing the Multiscale Progressive Mixing Chain (w/o MPMC) increases MSE by 2.29\% on average and 4.55\% at most, confirming that MPMC effectively captures multiscale dependencies for better TSF. It reduces memory overhead while enabling progressive fine-to-coarse interaction. Similar results are observed in the long-term TSF ablation with longer input and prediction length in Table~\ref{tab:ultra_long_ablation_2560_overall}.

\textbf{(2)} To assess MPMC’s progressive interaction, we removed it and directly concatenated multi-scale inputs. As shown in Figure~\ref{fig:mpmc_concat}, prediction errors increased, indicating that MPMC’s fine-to-coarse interaction helps capture multi-scale dependencies. \textbf{By addressing semantic gaps and noise between non-adjacent scales}, MPMC enables better alignment and integration of multi-scale information. Additional input-level results are provided in Appendix Table~\ref{tab:enhanced_scales}.

\begin{table*}[h!]
 % \vspace{-0.2cm}
    % \renewcommand{\arraystretch}{0.5}
    % \setlength{\tabcolsep}{2.75pt}
    % {|>{\setlength{\tabcolsep}{3pt}}c|c|c|}
    \centering
    \caption{Further ablation study and noise robustness experiments, reporting average results for forecasting 1020, 1320, and 1620 steps with input length 2560. Consistent gains are also observed with input length 2048. Lower MSE, MAE, and lower $\uparrow \%$ indicate better performance. Best results are in bold and worst in double underline.}
    \vspace{-0.1cm}
    \label{tab:ultra_long_ablation_2560_overall}
    % \begin{tabular}{c|c|p{20pt}p{20pt}|cc|cc|cc|cc|cc}
    {\small
    \begin{tabular}{c|c|cc|cc|cc|cc|cc}
        \hline
        \multirow{2}{*}{\shortstack{}} & &  \multicolumn{2}{c|}{ETTh1} &  \multicolumn{2}{c|}{ETTh2} & \multicolumn{2}{c|}{ETTm1} & \multicolumn{2}{c|}{ETTm2}& \multicolumn{2}{c}{Weather}
       \\
         & & MSE & MAE & MSE & MAE & MSE & MAE  & MSE & MAE & MSE & MAE\\
         \hline
\multirow{4}{*}{\rotatebox[origin=c]{0}{\shortstack{w/o \\noise}}}&SEMixer&\textbf{0.596} &\textbf{0.554}&\textbf{0.551} &\textbf{0.536}&\textbf{0.425} &\textbf{0.43}&\textbf{0.355} &\textbf{0.398}&\textbf{0.323} &\textbf{0.349}\\
\cline{2-12}
&w/o MPMC&\uuline{0.635} &0.577&\uuline{0.585} &\uuline{0.555}&0.43 &0.433&0.365 &0.405&0.327 &0.352\\
\cline{2-12}
&w/ SAM&0.638 &\uuline{0.579}&0.567 &0.546&\uuline{0.433} &\uuline{0.437}&0.359 &0.402&0.326 &0.351\\
\cline{2-12}
&w/o RAM&0.612 &0.562&0.568 &0.546&0.427 &0.432&\uuline{0.373} &\uuline{0.412}&\uuline{0.334} &\uuline{0.362}\\
\midrule[0.5pt]
\multirow{8}{*}{\rotatebox[origin=c]{0}{\shortstack{w/ noise \\($\varepsilon$=0.1)}}}&SEMixer&\textbf{0.619} &\textbf{0.569}&\textbf{0.605} &\textbf{0.568}&\textbf{0.443} &\textbf{0.445}&\textbf{0.427} &\textbf{0.452}&\textbf{0.325} &\textbf{0.352}\\
\cline{2-12}
&$\uparrow$\% Error&\textbf{$\uparrow$3.86\%} &\textbf{$\uparrow$2.71\%}&\textbf{$\uparrow$9.8\%} &\textbf{$\uparrow$5.97\%}&\textbf{$\uparrow$4.24\%} &\textbf{$\uparrow$3.49\%}&\textbf{$\uparrow$20.28\%} &\textbf{$\uparrow$13.57\%}&\textbf{$\uparrow$0.62\%} &\textbf{$\uparrow$0.86\%}\\
\cline{2-12}
&w/o MPMC&\uuline{0.693} &\uuline{0.611}&\uuline{0.715} &\uuline{0.62}&\uuline{0.464} &\uuline{0.463}&\uuline{0.502} &\uuline{0.497}&0.33 &0.356\\
\cline{2-12}
&$\uparrow$\% Error&\uuline{$\uparrow$9.13\%} &\uuline{$\uparrow$5.89\%}&\uuline{$\uparrow$22.22\%} &\uuline{$\uparrow$11.71\%}&\uuline{$\uparrow$7.91\%} &\uuline{$\uparrow$6.93\%}&\uuline{$\uparrow$37.53\%} &\uuline{$\uparrow$22.72\%}&$\uparrow$0.92\% &$\uparrow$1.14\%\\
\cline{2-12}
&w/ SAM&0.666 &0.595&0.635 &0.583&0.453 &0.454&0.465 &0.476&0.329 &0.355\\
\cline{2-12}
&$\uparrow$\% Error&$\uparrow$4.39\% &$\uparrow$2.76\%&$\uparrow$11.99\% &$\uparrow$6.78\%&$\uparrow$4.62\% &$\uparrow$3.89\%&$\uparrow$29.53\% &$\uparrow$18.41\%&$\uparrow$0.92\% &$\uparrow$1.14\%\\
\cline{2-12}
&w/o RAM&0.637 &0.577&0.627 &0.578&0.446 &0.448&0.449 &0.47&\uuline{0.346} &\uuline{0.374}\\
\cline{2-12}
&$\uparrow$\% Error&$\uparrow$4.08\% &$\uparrow$2.67\%&$\uparrow$10.39\% &$\uparrow$5.86\%&$\uparrow$4.45\% &$\uparrow$3.7\%&$\uparrow$20.38\% &$\uparrow$14.08\%&\uuline{$\uparrow$3.59\%} &\uuline{$\uparrow$3.31\%}\\
\midrule[0.5pt]
\multirow{8}{*}{\rotatebox[origin=c]{0}{\shortstack{w/ noise \\($\varepsilon$=0.3)}}}&SEMixer&\textbf{0.667} &\textbf{0.597}&\textbf{0.714} &\textbf{0.619}&\textbf{0.481} &\textbf{0.474}&\textbf{0.568} &\textbf{0.528}&\textbf{0.329} &\textbf{0.358}\\
\cline{2-12}
&$\uparrow$\% Error&\textbf{$\uparrow$11.91\%} &\textbf{$\uparrow$7.76\%}&\textbf{$\uparrow$29.58\%} &\textbf{$\uparrow$15.49\%}&\textbf{$\uparrow$13.18\%} &\textbf{$\uparrow$10.23\%}&\textbf{$\uparrow$60.0\%} &\textbf{$\uparrow$32.66\%}&\textbf{$\uparrow$1.86\%} &\textbf{$\uparrow$2.58\%}\\
\cline{2-12}
&w/o MPMC&\uuline{0.812} &\uuline{0.674}&\uuline{0.977} &\uuline{0.72}&\uuline{0.534} &\uuline{0.512}&\uuline{0.781} &\uuline{0.619}&0.336 &0.363\\
\cline{2-12}
&$\uparrow$\% Error&\uuline{$\uparrow$27.87\%} &\uuline{$\uparrow$16.81\%}&\uuline{$\uparrow$67.01\%} &\uuline{$\uparrow$29.73\%}&\uuline{$\uparrow$24.19\%} &\uuline{$\uparrow$18.24\%}&\uuline{$\uparrow$113.97\%} &\uuline{$\uparrow$52.84\%}&$\uparrow$2.75\% &$\uparrow$3.13\%\\
\cline{2-12}
&w/ SAM&0.731 &0.628&0.768 &0.642&0.494 &0.484&0.672 &0.576&0.336 &0.363\\
\cline{2-12}
&$\uparrow$\% Error&$\uparrow$14.58\% &$\uparrow$8.46\%&$\uparrow$35.45\% &$\uparrow$17.58\%&$\uparrow$14.09\% &$\uparrow$10.76\%&$\uparrow$87.19\% &$\uparrow$43.28\%&$\uparrow$3.07\% &$\uparrow$3.42\%\\
\cline{2-12}
&w/o RAM&0.689 &0.607&0.742 &0.631&0.483 &0.477&0.606 &0.551&\uuline{0.368} &\uuline{0.394}\\
\cline{2-12}
&$\uparrow$\% Error&$\uparrow$12.58\% &$\uparrow$8.01\%&$\uparrow$30.63\% &$\uparrow$15.57\%&$\uparrow$13.11\% &$\uparrow$10.42\%&$\uparrow$62.47\% &$\uparrow$33.74\%&\uuline{$\uparrow$10.18\%} &\uuline{$\uparrow$8.84\%}\\

        \hline

    \end{tabular}}
\vspace{-0.1cm}
\end{table*}

\subsubsection{\textbf{Demonstrating the effectiveness of RAM and MPMC through patch embedding
visualization}}

In our setting, the patch count ratios from the finest to the coarsest scales are $N^1:N^2:N^3:N^4=8:4:2:1$. 
After training, we extract the 120 learned patch embeddings from each test sample before the final linear predictor.
We randomly select 100 test samples, yielding a total of 12,000 patch embeddings, which are visualized using t-SNE~\cite{van2008visualizing} to further illustrate the effectiveness of RAM and MPMC.

As shown in Figure~\ref{fig:vis_tsne}(b), only using MPMC (without RAM), SEMixer learns four well-separated scale clusters, whereas removing the MPMC architecture results in unclear patterns (Figure~\ref{fig:vis_tsne}(c,d)), verifying the effectiveness of MPMC in multiscale representation learning.
In Figure~\ref{fig:vis_tsne}(e), further removing RAM after disabling MPMC further degrades representations, highlighting RAM’s effectiveness.

More importantly, when RAM is enabled together with MPMC, as shown in Figure~\ref{fig:vis_tsne}(a), the scale clusters become distinct yet well integrated, indicating that cross-scale interactions allow each scale to exploit information from others, thereby enhancing representations and achieving effective multiscale fusion. This joint representation enables SEMixer to better capture multiscale dependencies and yields higher forecasting accuracy.

\subsection{{Hyperparameter sensitivity of RAM}}

Figure~\ref{fig:Hyperparameter} analyzes the sensitivity of SEMixer to the random disconnection probability $p$ across different datasets.
In RAM, $p$ controls how many patch-to-patch interactions are randomly disconnected during training.
When $p$ is small, most patch interactions are preserved, but the sampled interaction patterns become less diverse, limiting the dropout-ensemble effect of RAM.
When $p$ is too large, the sampled interaction graphs become highly sparse, which increases interaction diversity but may remove too many useful connections and weaken semantic propagation among related patches.
Overall, SEMixer remains relatively stable over a broad range of $p$, suggesting that RAM is not overly sensitive to this hyperparameter.
Empirically, values around $p=0.85$ tend to provide strong performance across datasets, reflecting a practical trade-off between interaction diversity and semantic information preservation.
Therefore, we set $p$ to 0.85 in all experiments.

% Figure~\ref{fig:Hyperparameter} analyzes the sensitivity of SEMixer to the random disconnection probability $p$ across different datasets.
% The parameter $p$ controls the sparsity of patch interactions in RAM: a smaller $p$ preserves more connections but reduces the diversity of randomly sampled interaction patterns, while a larger $p$ encourages more diverse subgraphs but may weaken useful semantic propagation among patches.
% Overall, SEMixer remains relatively stable over a broad range of $p$, suggesting that RAM is not overly sensitive to this hyperparameter.
% Empirically, values around $p=0.85$ tend to provide strong performance across datasets, reflecting a practical trade-off between interaction diversity and semantic information preservation.
% Therefore, we set $p$ to 0.85 in all experiments.

\subsection{Noise robustness of SEMixer}
Firstly, Table~\ref{tab:ultra_long_ablation_2560_overall} further validates the effectiveness of the proposed MPMC and RAM under longer input and prediction lengths. \textbf{Removing either MPMC or RAM leads to a significant increase in prediction error}. 
This improvement benefits from the  design of MPMC, which facilitates more effective multi-scale representation learning, while RAM enriches the semantic information of each time patch through random interactions, enabling MPMC to better capture multi-scale temporal dependencies.

Secondly, time series often contain noise, so robustness is essential.
We evaluate SEMixer by injecting noise of magnitude $\varepsilon \times 100\%$ within $[-2\mathcal{X}, 2\mathcal{X}]$ into the test samples. This setting simulates heteroscedastic noise, where the noise intensity depends on the original signal amplitude, reflecting real-world scenarios in which noise is correlated with signal strength.

As shown in the orange-highlighted text of Table~\ref{tab:ultra_long_ablation_2560_overall}, removing MPMC greatly increases error, indicating its key role in noise resistance. 
This is because real-world signals often exhibit cross-scale consistency (e.g., shared trends).
MPMC amplifies such consistent features while suppressing noise, enabling stable performance under noisy conditions.

Compared to standard SAM, our RAM is more noise-resistant because it learns diverse interactions during training, reducing the impact of noisy ones.
In contrast, SAM’s limited heads make it prone to overfitting to noise.
This further demonstrates RAM’s effectiveness in enhancing performance, reducing overhead, and improving robustness.

\section{Conclusion}
In this paper, we propose SEMixer, a lightweight multiscale model designed to capture temporal dependencies across multiscale inputs for accurate and efficient long-term TSF.
SEMixer is built on two core components: the Random Attention Mechanism (RAM) and the Multiscale Progressive Mixing Chain (MPMC).
RAM enhances patch semantics by learning numerous and diverse time-patch interactions during training and integrating them via a dropout ensemble at inference, improving the effectiveness of MLP-Mixer.
MPMC explicitly addresses computational overhead and semantic noise across scales, enabling efficient and robust multiscale representation learning, while the semantic enhancement provided by RAM further helps MPMC capture multi-scale temporal dependencies.
Extensive experiments show that SEMixer consistently outperforms transformer-, CNN-, and linear-based TSF baselines on ten public datasets and the \textit{2025 CCF AlOps Challenge}.

% In this paper, we propose SEMixer, a lightweight multiscale model designed to capture temporal dependencies across multiscale inputs for accurate and efficient long-term TSF.
% SEMixer is built upon two core components: the Random Attention Mechanism (RAM) and the Multiscale Progressive Mixing Chain (MPMC). 
% RAM learns a large number of time-patch interaction patterns during training and integrates them via a dropout ensemble during inference, thereby enhancing patch semantics and improving the effectiveness of MLP-Mixer. 
% MPMC considers the computational overhead and semantic noise in multi-scale inputs, endowing the model with efficient and robust multi-scale representation learning capability. 
% The semantic enhancement provided by RAM helps MPMC better capture multi-scale temporal dependencies.
% Extensive experimental results demonstrate that SEMixer consistently outperforms popular TSF baselines, including transformer-, CNN-, and linear-based models.
% These advantages are validated not only on ten widely used public datasets but also in the \textit{2025 CCF AlOps Challenge}.

\section{Acknowledgements}
This work was supported by the National Natural Science Foundation of China under Grant 62427819.


\bibliographystyle{ACM-Reference-Format}
\balance
\bibliography{sample-base}

\appendix

\section{Appendices}

\subsection{Datasets}

The 10 public datasets used in this paper are extensively used for long-term TSF algorithm evaluation, 
covering multiple fields including industry (4 ETT datasets), climate (weather), energy (Solar Energy and Electricity), health (ILI), and economy (Exchange). The detailed descriptions are as follows:
\begin{enumerate}
\item Electricity dataset\footnote{\url{https://archive.ics.uci.edu/dataset/321/electricity}} collects the electricity consumption (kWh) every 15 minutes of 321 clients from 2012 to 2014.
\item ETT datasets\footnote{\url{https://github.com/zhouhaoyi/Informer2020}} comprises two sub-datasets, ETT1 and ETT2, collected from two separate counties. Each sub-dataset offers two versions with varying sampling resolutions (15 minutes and 1 hour). ETT dataset includes multiple time series of electrical loads and a single time sequence of oil temperature.

\item Weather dataset\footnote{\url{https://www.bgc-jena.mpg.de/wetter/}} contains 21 meteorological indicators, such as air temperature, humidity, etc, recorded every 10 minutes for the entirety of 2020.
\item Exchange dataset\footnote{\url{https://github.com/laiguokun/multivariate-time-series-data}} contains the current exchange of eight countries.
\item ILI(Influenza-Like Illness)\footnote{\url{https://gis.cdc.gov/grasp/fluview/fluportaldashboard.html}} dataset contains the influenza-like illness patients in the United States.
\item Solar-Energy\footnote{\url{https://github.com/laiguokun/multivariate-time-series-data}} records the solar power production of 137 PV plants in 2006,
which are sampled every 10 minutes.

\item Traffic records hourly road occupancy rates measured by 862 sensors of the San Francisco Bay area freeways in 2 years. 

\end{enumerate}

% \subsubsection{Datasets}
\begin{table}[h!] %[h!]
    \setlength{\tabcolsep}{3.5pt}
    % {|>{\setlength{\tabcolsep}{3pt}}c|c|c|}
    \centering
    % \vspace{-0.2cm}
    \caption{Statistics of ten public datasets. \textit{Data size} denotes the number of samples in the train, validation, and test sets.  \textit{Frequency} denotes the sampling interval of time points.}
    % \vspace{-0.3cm}
    \label{tab:dataset_stat}
    % \begin{tabular}{c|c|p{20pt}p{20pt}|cc|cc|cc|cc|cc}
    { \small
    \begin{tabular}{c|c|c|c}
        \hline
        \multirow{1}{*}{\shortstack{Datasets}}  &  \multicolumn{1}{c|}{Variable} &  \multicolumn{1}{c|}{Data size }  & \multicolumn{1}{c}{Frequency} \\
         \midrule[0.5pt]
         \multirow{1}{*}{ETTh1,ETTh2}  &7 &\shortstack{(8545, 2881, 2881)} &Hourly\\
        \midrule[0.5pt]
         \multirow{1}{*}{ETTm1,ETTm2}  &7 &\shortstack{(34465, 11521, 11521)}  &15min\\
        \midrule[0.5pt]
        \multirow{1}{*}{Weather}  &21 &\shortstack{(36792, 5271, 10540)} &10min\\
        \midrule[0.5pt]
        \multirow{1}{*}{Exchange rate}  &9 &\shortstack{(5120, 665, 1422)} &Daily\\
                \midrule[0.5pt]
        \multirow{1}{*}{Electricity }  &321 &\shortstack{(18317, 2633, 5261)}&Hourly\\
                \midrule[0.5pt]
        \multirow{1}{*}{Solar energy }  &137 &\shortstack{(36601, 5161, 10417)}&10min\\
        \midrule[0.5pt]
        \multirow{1}{*}{ILI (Influenza-Like Illness) }  &7 &\shortstack{(617, 74, 170)}&Daily\\
                \midrule[0.5pt]
        \multirow{1}{*}{Traffic }  &862 &\shortstack{(12185, 1757, 3509)}&Hourly\\
        \midrule[0.5pt]
    \end{tabular}}
    % } % \small
% \vspace{-0.5cm}
\end{table}

\subsection{Supplementary experimental results}

\subsubsection{More ablation studies on MPMC structure}
The results are shown in Table~\ref{tab:enhanced_scales}. 
The MSE of $\mathcal{{X}}_d^{s}$ is obtained by directly sending $\mathcal{{X}}_d^{s}$ to the linear predictor while the MSE of $\mathcal{\bar{X}}_d^{s}$ is obtained by first sending $\mathcal{{X}}_d^{s}$ to a Temporal Mixing Block and then forwarding the output to the linear predictor. 
MSE of $\mathcal{\bar{X}}_{d_{*}}^{s}$ is obtained by first concatenating $\mathcal{\bar{X}}_{d_{*}}^{s-1}$ and $\mathcal{{X}}_d^{s}$, sending the concatenation into a Temporal Mixing Block for multi-scale mixing (MPMC structure) and then separating $\mathcal{\bar{X}}_{d_{*}}^{s}$ from the output and forwarding it to the linear predictor. 

From Table~\ref{tab:enhanced_scales} we can observe that at different scale inputs, $\mathcal{{X}}_d^{s}$ is the worst because it lacks any interactions within the scale input, i.e., intra-scale interactions. The second worst is $\mathcal{\bar{X}}_d^{s}$ because it lacks interactions across multi-scale inputs (inter-scale interactions). 
$\mathcal{\bar{X}}_{d_{*}}^{s}$ using the MPMC achieves the best because it includes both interactions within a single-scale input and across multi-scale inputs. 
This indicates that the proposed MPMC structure is reasonable. 
It enhances the representation capability of different-scale inputs and improves the TSF accuracy.
\begin{table}[h!]
% \vspace{-0.2cm}
    % \renewcommand{\arraystretch}{0.5}
    \setlength{\tabcolsep}{2.5pt}
    % {|>{\setlength{\tabcolsep}{3pt}}c|c|c|}
    \centering
    \caption{MSE of using different scale inputs to predict future 96 time steps under different multi-scale structures. }
    \label{tab:enhanced_scales}
    \vspace{-0.2cm}
    % \begin{tabular}{c|c|p{20pt}p{20pt}|cc|cc|cc|cc|cc}
    {\small
    \begin{tabular}{c|c|c|c|c|c|c|c|c}
        \hline
        &   \multicolumn{1}{c|}{Ettm1} &  \multicolumn{1}{c|}{Weather}&   & \multicolumn{1}{c|}{Ettm1} & \multicolumn{1}{c|}{Weather}&& \multicolumn{1}{c|}{Ettm1} & \multicolumn{1}{c}{Weather}\\
       
         \midrule[1pt]
         \multirow{1}{*}{$\mathcal{\bar{X}}_{d_{*}}^{2}$}  
           &\textbf{0.295}&\textbf{0.149} &\multirow{1}{*}{$\mathcal{\bar{X}}_{d_{*}}^{3}$}&\textbf{0.296}&\textbf{0.149}&\multirow{1}{*}{$\mathcal{\bar{X}}_{d_{*}}^{4}$}&\textbf{0.297}&\textbf{0.149}\\
        \midrule[1pt]
        \multirow{1}{*}{$\mathcal{\bar{X}}_d^{2}$}  
           &0.297&0.152&\multirow{1}{*}{$\mathcal{\bar{X}}_d^{3}$}&0.298&0.153&\multirow{1}{*}{$\mathcal{\bar{X}}_d^{4}$}&0.297&0.159 \\
        \midrule[1pt]
        \multirow{1}{*}{$\mathcal{{X}}_d^{2}$}  
           &0.304&0.169&\multirow{1}{*}{$\mathcal{{X}}_d^{3}$}&0.304&0.168&\multirow{1}{*}{$\mathcal{{X}}_d^{4}$}&0.304&0.168 \\
        \midrule[1pt]
    \end{tabular}}
\vspace{-0.4cm}
\end{table}

\subsubsection{Time complexity analysis of random attention mechanism (RAM)}
\label{sec:time_complexity}
Assuming the number of time patches and the layer hidden size are $N$ and $H$, respectively, the time complexity of RAM consists of $\mathcal{O}(N^2)$ for randomly sampling binary 0--1 matrices and $\mathcal{O}(N^{2}H)$ for patch interaction by matrix multiplication. 
In contrast, the time complexity of the self-attention mechanism (SAM) is $\mathcal{O}(N^{2}H) + \mathcal{O}(N^2) + \mathcal{O}(N^2) + \mathcal{O}(N^{2}H)$, corresponding to query-key dot-product computation, rescaling, softmax normalization, and patch interaction, respectively. 
Hence, the time complexity of RAM is lower than that of SAM. 
We also calculate the FLOPs of different models using the \textit{fvcore} library in PyTorch with input size $[1,1024,321]$. 
The FLOPs of PatchTST, TimeMixer, TSMixer, and SEMixer are $2.85 \times 10^{12}$, $5.68 \times 10^{10}$, $8.55 \times 10^{9}$, and $4.67 \times 10^{9}$, respectively.

\subsection{Supplemental results }
In Table~\ref{tab:metric_public_long_term} and Table~\ref{tab:metric_public_ultra_long_term} of the main text, we report the average prediction error of different baselines across all prediction lengths; here, we provide the results for each individual prediction length.

\subsubsection{Full results of long-term forecasting (Table~\ref{tab:metric_public_long_term_appendix_p1}, Table~\ref{tab:ultra_long_appendix_2560}, Table~\ref{tab:metric_public_long_term_appendix_p2} and Table~\ref{tab:ultra_long_appendix_2048})}

We observe that some models perform well with shorter input lengths (e.g., 96) but degrade with longer input lengths, becoming weaker than other baselines. To avoid potential unfair comparisons, besides evaluating with a fixed input length, we also search for the optimal input length from {96, 384, 512, 640, 768, 1024, 1280, 1536, 1664, 1792, 2048} for each baseline and report their best results. The forecasting results for 96, 192, 336, and 720 steps are presented in Table~\ref{tab:metric_public_long_term_appendix_p1} and Table~\ref{tab:metric_public_long_term_appendix_p2}.

The results for forecasting of 1020, 1320, and 1620 steps with fixed input lengths of 2560 and 2048 are shown in Table~\ref{tab:ultra_long_appendix_2560} and Table~\ref{tab:ultra_long_appendix_2048}. 

Overall, SEMixer consistently achieves the best performance under both optimal input length searching and fixed input length settings.

\begin{table*}[h]
 % \vspace{-0.2cm}
    % \renewcommand{\arraystretch}{0.5}
    \setlength{\tabcolsep}{6.5pt}
    % {|>{\setlength{\tabcolsep}{3pt}}c|c|c|}
    \centering
    \caption{Full results for forecasting horizons of 96, 192, 336, and 720 steps (Part 1).
The input length is selected from \{96, 384, 512, 640, 768, 1024, 1280, 1536, 1664, 1792, 2048\}.}
    % \vspace{-0.2cm}
    \label{tab:metric_public_long_term_appendix_p1}
    % \begin{tabular}{c|c|p{20pt}p{20pt}|cc|cc|cc|cc|cc}
    {\footnotesize
    \begin{tabular}{c|c|cc|cc|cc|cc|cc|cc|cc}
        \hline
        \multirow{2}{*}{\shortstack{}} & &  \multicolumn{2}{c|}{SEMixer (Ours)} &  \multicolumn{2}{c|}{DeformableTST} & \multicolumn{2}{c|}{TimeXer} & \multicolumn{2}{c|}{ModernTCN}& \multicolumn{2}{c|}{Pathformer}
        &  \multicolumn{2}{c|}{iTransformer}
        & \multicolumn{2}{c}{TimesNet}\\
         & & MSE & MAE & MSE & MAE & MSE & MAE  & MSE & MAE & MSE & MAE& MSE & MAE& MSE & MAE\\ 
         \midrule[0.5pt]
\multirow{4}{*}{\rotatebox[origin=c]{90}{ETTh1}}&96&\textbf{0.365}$\pm$2e-4 &\textbf{0.394}$\pm$2e-4&\underline{0.368} &\underline{0.402}&0.385 &0.41&0.379 &0.41&0.393 &0.406&0.399 &0.425&0.384 &0.402\\
&192&\textbf{0.398}$\pm$5e-4 &\textbf{0.417}$\pm$3e-4&\underline{0.402} &\underline{0.424}&0.415 &0.433&0.415 &0.437&0.421 &0.436&0.418 &0.441&0.436 &0.429\\
&336&\underline{0.421}$\pm$6e-4 &\textbf{0.422}$\pm$3e-4&\textbf{0.419} &\underline{0.433}&0.43 &0.446&0.438 &0.451&0.451 &0.451&0.443 &0.459&0.491 &0.469\\
&720&\textbf{0.417}$\pm$7e-3 &\textbf{0.439}$\pm$4e-3&\underline{0.443} &\underline{0.457}&0.471 &0.486&0.474 &0.481&0.483 &0.47&0.495 &0.496&0.512 &0.494\\
\midrule[0.5pt]
\multirow{4}{*}{\rotatebox[origin=c]{90}{ETTh2}}&96&\textbf{0.269}$\pm$5e-4 &\textbf{0.336}$\pm$3e-4&0.298 &0.349&0.282 &0.344&\underline{0.272} &\underline{0.34}&0.285 &0.349&0.295 &0.356&0.315 &0.362\\
&192&\textbf{0.325}$\pm$9e-4 &\textbf{0.372}$\pm$6e-4&0.356 &0.388&0.344 &0.386&\underline{0.33} &\underline{0.381}&0.331 &0.385&0.36 &0.397&0.402 &0.414\\
&336&\textbf{0.348}$\pm$5e-4 &\textbf{0.394}$\pm$4e-4&0.379 &0.415&0.369 &0.407&\underline{0.358} &\underline{0.407}&0.368 &0.409&0.392 &0.42&0.441 &0.457\\
&720&\textbf{0.381}$\pm$8e-4 &\textbf{0.426}$\pm$9e-4&0.42 &0.457&{0.386} &\underline{0.426}&\underline{0.382} &0.434&0.389 &0.427&0.415 &0.445&0.446 &0.469\\
\midrule[0.5pt]
\multirow{4}{*}{\rotatebox[origin=c]{90}{ETTm1}}&96&\textbf{0.291}$\pm$2e-3 &\textbf{0.346}$\pm$2e-3&\underline{0.291} &\underline{0.348}&0.31 &0.36&0.302 &0.355&0.301 &0.352&0.303 &0.358&0.338 &0.375\\
&192&\textbf{0.329}$\pm$8e-4 &\textbf{0.364}$\pm$2e-3&\underline{0.334} &\underline{0.373}&0.352 &0.387&0.346 &0.378&0.356 &0.383&0.34 &0.379&0.363 &0.386\\
&336&\textbf{0.354}$\pm$1e-3 &\textbf{0.384}$\pm$9e-4&\underline{0.363} &\underline{0.39}&0.378 &0.405&0.375 &0.404&0.387 &0.405&0.372 &0.404&0.389 &0.405\\
&720&\textbf{0.393}$\pm$6e-4 &\textbf{0.407}$\pm$3e-4&\underline{0.416} &\underline{0.42}&0.425 &0.428&0.417 &0.421&0.416 &0.42&0.431 &0.444&0.478 &0.45\\
\midrule[0.5pt]
\multirow{4}{*}{\rotatebox[origin=c]{90}{ETTm2}}&96&\textbf{0.161}$\pm$2e-4 &\textbf{0.254}$\pm$5e-4&0.175 &0.264&0.17 &0.26&0.166 &0.264&\underline{0.168} &\underline{0.258}&0.178 &0.271&0.187 &0.267\\
&192&\textbf{0.213}$\pm$2e-3 &\textbf{0.295}$\pm$2e-3&0.234 &0.3&0.233 &0.315&\underline{0.22} &\underline{0.299}&0.227 &0.299&0.233 &0.315&0.249 &0.309\\
&336&\textbf{0.259}$\pm$1e-3 &\textbf{0.326}$\pm$1e-3&0.283 &0.332&0.282 &0.333&0.293 &0.342&\underline{0.273} &\underline{0.331}&0.282 &0.345&0.312 &0.358\\
&720&\textbf{0.33}$\pm$3e-3 &\textbf{0.373}$\pm$6e-4&0.361 &0.39&0.366 &0.391&0.355 &0.392&0.366 &0.392&\underline{0.354} &\underline{0.391}&0.408 &0.403\\
\midrule[0.5pt]
\multirow{4}{*}{\rotatebox[origin=c]{90}{Weather}}&96&\textbf{0.145}$\pm$3e-3 &\textbf{0.194}$\pm$2e-3&\underline{0.146} &\underline{0.195}&0.152 &0.203&0.146 &0.201&0.155 &0.208&0.16 &0.212&0.161 &0.216\\
&192&\textbf{0.188}$\pm$4e-3 &\textbf{0.236}$\pm$4e-3&\underline{0.191} &\underline{0.238}&0.194 &0.243&0.194 &0.246&0.196 &0.246&0.205 &0.252&0.219 &0.261\\
&336&\textbf{0.235}$\pm$2e-3 &\textbf{0.278}$\pm$8e-4&\underline{0.241} &\underline{0.278}&0.247 &0.285&0.245 &0.285&0.25 &0.286&0.255 &0.291&0.28 &0.306\\
&720&\textbf{0.296}$\pm$1e-3 &\textbf{0.326}$\pm$9e-4&\underline{0.305} &\underline{0.331}&0.31 &0.334&0.312 &0.333&0.324 &0.337&0.322 &0.337&0.365 &0.359\\
\midrule[0.5pt]
\multirow{4}{*}{\rotatebox[origin=c]{90}{Electricity}}&96&\textbf{0.127}$\pm$3e-4 &\textbf{0.222}$\pm$2e-4&0.132 &0.234&0.14 &0.246&\underline{0.129} &\underline{0.226}&0.134 &0.236&0.142 &0.242&0.168 &0.272\\
&192&\textbf{0.143}$\pm$8e-5 &\textbf{0.238}$\pm$6e-5&0.148 &0.248&0.158 &0.263&\underline{0.143} &\underline{0.239}&0.156 &0.256&0.159 &0.259&0.184 &0.289\\
&336&\textbf{0.157}$\pm$6e-4 &\textbf{0.254}$\pm$2e-4&0.165 &0.266&0.169 &0.274&\underline{0.161} &\underline{0.259}&0.179 &0.271&0.167 &0.269&0.198 &0.3\\
&720&\textbf{0.188}$\pm$3e-4 &\textbf{0.283}$\pm$5e-4&0.197 &0.296&0.19 &0.292&\underline{0.191} &\underline{0.286}&0.209 &0.307&0.192 &0.293&0.22 &0.32\\
\midrule[0.5pt]
\multirow{4}{*}{\rotatebox[origin=c]{90}{ILl}}&96&\textbf{2.411}$\pm$0.012 &\textbf{1.085}$\pm$0.005&2.757 &1.164&2.72 &1.167&2.828 &1.204&2.68 &1.12&\underline{2.563} &\underline{1.098}&3.73 &1.28\\
&192&\textbf{2.324}$\pm$0.057 &\textbf{1.06}$\pm$0.015&2.717 &1.139&2.661 &1.151&2.768 &1.18&2.8 &1.16&\underline{2.597} &\underline{1.112}&3.13 &1.16\\
&336&\textbf{2.352}$\pm$0.062 &\textbf{1.084}$\pm$0.003&2.824 &1.179&2.745 &1.15&2.872 &1.191&2.7 &1.14&\underline{2.559} &\underline{1.102}&3.38 &1.22\\
&720&\textbf{2.453}$\pm$0.064 &\textbf{1.091}$\pm$0.015&2.899 &1.214&2.907 &1.176&3.122 &1.225&2.83 &1.16&\underline{2.621} &\underline{1.133}&4.15 &1.37\\
\midrule[0.5pt]
\multirow{4}{*}{\rotatebox[origin=c]{90}{Exchange}}&96&\textbf{0.086}$\pm$8e-4 &\textbf{0.206}$\pm$0.001&\underline{0.097} &\underline{0.22}&0.099 &0.223&0.101 &0.226&0.111 &0.237&0.112 &0.241&0.231 &0.36\\
&192&\textbf{0.176}$\pm$0.001 &\textbf{0.301}$\pm$0.002&\underline{0.196} &\underline{0.317}&0.208 &0.325&0.258 &0.375&0.218 &0.333&0.204 &0.328&0.433 &0.494\\
&336&\textbf{0.322}$\pm$0.007 &\textbf{0.408}$\pm$0.009&0.366 &0.441&0.385 &0.451&0.401 &0.454&0.434 &0.476&\underline{0.359} &\underline{0.437}&0.714 &0.621\\
&720&\textbf{0.793}$\pm$0.04 &\textbf{0.674}$\pm$0.02 &1.032 &0.76&0.852 &0.675&1.203 &0.801&1.439 &0.872&\underline{0.806} &\underline{0.697}&1.03 &0.747\\
\midrule[0.5pt]
\multirow{4}{*}{\rotatebox[origin=c]{90}{Solar Energy}}&96&\textbf{0.167}$\pm$0.003 &\textbf{0.229}$\pm$9e-4&\underline{0.171} &\underline{0.225}&0.176 &0.26&0.197 &0.28&0.201 &0.253&0.198 &0.273&0.219 &0.279\\
&192&\textbf{0.179}$\pm$5e-4 &\textbf{0.242}$\pm$0.002&\underline{0.181} &\underline{0.24}&0.186 &0.27&0.216 &0.299&0.235 &0.275&0.229 &0.297&0.228 &0.297\\
&336&\underline{0.188}$\pm$0.001 &\underline{0.249}$\pm$0.001&\textbf{0.186} &\textbf{0.243}&0.194 &0.264&0.235 &0.316&0.263 &0.308&0.246 &0.31&0.239 &0.312\\
&720&\underline{0.199}$\pm$3e-4 &\underline{0.259}$\pm$0.001&\textbf{0.198} &\textbf{0.257}&0.204 &0.276&0.233 &0.329&0.262 &0.297&0.254 &0.318&0.238 &0.313\\
\midrule[0.5pt]
\multirow{4}{*}{\rotatebox[origin=c]{90}{Traffic}}&96&\underline{0.361}$\pm$3e-4 &\textbf{0.256}$\pm$4e-4&\textbf{0.36} &\underline{0.261}&0.368 &0.271&0.366 &0.270&0.389 &0.277&0.38 &0.291&0.556 &0.305\\
&192&\underline{0.379}$\pm$8e-4 &\underline{0.263}$\pm$6e-4&0.383 &0.267&0.38 &0.275&\textbf{0.379} &\textbf{0.261}&0.396 &0.283&0.4 &0.306&0.557 &0.304\\
&336&\textbf{0.390}$\pm$5e-4 &\textbf{0.269}$\pm$7e-4&\underline{0.393} &0.281&0.402 &0.286&{0.395} &\underline{0.279}&0.417 &0.307&0.42 &0.317&0.573 &0.304\\
&720&\textbf{0.421}$\pm$3e-5 &\textbf{0.283}$\pm$6e-4&\underline{0.435} &{0.3}&0.44 &0.302&0.437 &\underline{0.297}&0.447 &0.319&0.466 &0.344&0.601 &0.318\\
\hline

        % \midrule[0.5pt]
        % \hline
    \end{tabular}}
% \vspace{-0.4cm}
\end{table*}

\begin{table*}[h]
 % \vspace{-0.2cm}
    % \renewcommand{\arraystretch}{0.5}
    \setlength{\tabcolsep}{2.5pt}
    % {|>{\setlength{\tabcolsep}{3pt}}c|c|c|}
    \centering
    \caption{Full results of forecasting longer horizons 1020, 1320, and 1620 with input lengths 2048. ``-'' denotes out of memory.}
    % \vspace{-0.2cm}
    \label{tab:ultra_long_appendix_2560}
    % \begin{tabular}{c|c|p{20pt}p{20pt}|cc|cc|cc|cc|cc}
    {\footnotesize
    \begin{tabular}{c|c|cc|cc|cc|cc|cc|cc|cc|cc|cc|cc|cc}
        \hline
        \multirow{2}{*}{\shortstack{}} & &  \multicolumn{2}{c|}{SEMixer} &  \multicolumn{2}{c|}{DeformableTST} & \multicolumn{2}{c|}{TimeXer} & \multicolumn{2}{c|}{ModernTCN}& \multicolumn{2}{c|}{iTransformer}
        &  \multicolumn{2}{c|}{TimesNet}
        & \multicolumn{2}{c|}{TSMixer}& \multicolumn{2}{c|}{DLinear}& \multicolumn{2}{c|}{PatchTST}& \multicolumn{2}{c|}{TimeMixer}& \multicolumn{2}{c}{Scaleformer}\\
         & & MSE & MAE & MSE & MAE & MSE & MAE  & MSE & MAE & MSE & MAE& MSE & MAE& MSE & MAE& MSE & MAE& MSE & MAE& MSE & MAE& MSE & MAE\\ 
         \midrule[0.5pt]
\multirow{3}{*}{\rotatebox[origin=c]{90}{ETTh1}}&1020&\text{0.533} &\text{0.52}&0.623 &0.566&0.636 &0.595&0.709 &0.59&0.71 &0.628&1.089 &0.837&\underline{0.552} &\underline{0.533}&0.634 &0.59&0.565 &0.539&0.799 &0.638&0.736 &0.606\\
&1320&\text{0.593} &\text{0.552}&0.73 &0.618&0.732 &0.64&0.874 &0.673&0.848 &0.709&1.153 &0.838&\underline{0.625} &\underline{0.568}&0.742 &0.644&0.626 &0.569&0.874 &0.685&0.993 &0.758\\
&1620&\text{0.661} &\text{0.591}&0.83 &0.651&0.888 &0.709&1.049 &0.81&1.011 &0.784&1.435 &0.948&0.721 &0.616&0.858 &0.701&\underline{0.69} &\underline{0.604}&1.127 &0.731&0.997 &0.775\\
\midrule[0.5pt]
\multirow{3}{*}{\rotatebox[origin=c]{90}{ETTh2}}&1020&\underline{0.496} &\underline{0.508}&0.564 &0.546&0.573 &0.545&0.666 &0.577&0.602 &0.568&0.655 &0.6&\text{0.49} &\text{0.51}&1.288 &0.774&0.5 &0.515&0.52 &0.527&0.548 &0.552\\
&1320&\text{0.536} &\text{0.53}&0.596 &0.571&0.606 &0.574&0.905 &0.716&0.669 &0.603&0.706 &0.617&0.561 &0.548&1.407 &0.816&\underline{0.545} &\underline{0.543}&0.58 &0.564&0.572 &0.564\\
&1620&0.62 &0.57&0.689 &0.613&\underline{0.625} &\underline{0.562}&0.969 &0.732&0.632 &0.582&0.999 &0.732&0.643 &0.583&1.443 &0.838&\text{0.597} &\text{0.565}&0.744 &0.635&0.653 &0.597\\
\midrule[0.5pt]
\multirow{3}{*}{\rotatebox[origin=c]{90}{ETTm1}}&1020&\text{0.415} &\text{0.422}&0.435 &0.44&0.448 &0.46&0.445 &0.457&0.47 &0.475&- &-&0.422 &0.435&\underline{0.419} &\underline{0.436}&0.44 &0.446&0.462 &0.464&0.433 &0.444\\
&1320&\text{0.427} &\text{0.431}&0.454 &0.461&0.456 &0.468&0.463 &0.467&0.486 &0.483&- &-&0.435 &0.445&\underline{0.427} &\underline{0.449}&0.449 &0.453&0.536 &0.505&0.446 &0.456\\
&1620&\text{0.433} &\text{0.437}&0.454 &0.45&0.458 &0.47&0.481 &0.475&0.511 &0.499&- &-&\underline{0.435} &\underline{0.444}&0.439 &0.463&0.451 &0.456&0.519 &0.501&0.451 &0.459\\
\midrule[0.5pt]
\multirow{3}{*}{\rotatebox[origin=c]{90}{ETTm2}}&1020&\text{0.353} &\text{0.393}&0.378 &0.409&0.376 &0.409&0.398 &0.428&0.4 &0.424&- &-&0.372 &0.409&0.44 &0.454&\underline{0.368} &\underline{0.404}&0.42 &0.432&0.387 &0.425\\
&1320&\text{0.36} &\text{0.401}&0.38 &0.416&0.376 &0.416&0.401 &0.437&0.431 &0.443&- &-&0.376 &0.416&0.49 &0.484&\underline{0.374} &\underline{0.411}&0.419 &0.44&0.39 &0.429\\
&1620&\text{0.353} &\text{0.399}&0.38 &0.421&0.371 &0.414&0.397 &0.437&0.424 &0.443&- &-&0.372 &0.418&0.503 &0.494&\underline{0.367} &\underline{0.412}&0.429 &0.446&0.386 &0.429\\
\midrule[0.5pt]
\multirow{3}{*}{\rotatebox[origin=c]{90}{Weather}}&1020&\text{0.312} &\text{0.342}&0.325 &0.353&0.324 &0.354&0.344 &0.374&0.367 &0.39&- &-&0.33 &0.36&0.318 &0.358&\underline{0.32} &\underline{0.351}&- &-&0.329 &0.365\\
&1320&\text{0.323} &\text{0.349}&0.336 &0.358&0.334 &0.361&0.352 &0.379&0.342 &0.364&- &-&0.337 &0.365&0.33 &0.367&\underline{0.33} &\underline{0.357}&- &-&0.338 &0.368\\
&1620&\text{0.333} &\text{0.357}&0.344 &0.367&0.342 &0.367&0.36 &0.385&0.349 &0.372&- &-&0.346 &0.371&0.34 &0.374&\underline{0.341} &\underline{0.365}&- &-&0.348 &0.38\\
\hline

    \end{tabular}}
% \vspace{-0.4cm}
\end{table*}

\begin{table*}[h]
 % \vspace{-0.2cm}
    % \renewcommand{\arraystretch}{0.5}
    \setlength{\tabcolsep}{6.75pt}
    % {|>{\setlength{\tabcolsep}{3pt}}c|c|c|}
    \centering
    \caption{Full results for forecasting horizons of 96, 192, 336, and 720 steps (Part 2).
The input length is selected from \{96, 384, 512, 640, 768, 1024, 1280, 1536, 1664, 1792, 2048\}. }
    % \vspace{-0.2cm}
    \label{tab:metric_public_long_term_appendix_p2}
    % \begin{tabular}{c|c|p{20pt}p{20pt}|cc|cc|cc|cc|cc}
    {\footnotesize
    \begin{tabular}{c|c|cc|cc|cc|cc|cc|cc|cc}
        \hline
        \multirow{2}{*}{\shortstack{}} & &  \multicolumn{2}{c|}{SEMixer (Ours)} &  \multicolumn{2}{c|}{TSMixer} & \multicolumn{2}{c|}{DLinear} & \multicolumn{2}{c|}{PatchTST}& \multicolumn{2}{c|}{TimeMixer}
        &  \multicolumn{2}{c|}{FiLM}
        & \multicolumn{2}{c}{Scaleformer}\\
         & & MSE & MAE & MSE & MAE & MSE & MAE  & MSE & MAE & MSE & MAE& MSE & MAE& MSE & MAE\\ 
         \midrule[0.5pt]
         \multirow{4}{*}{\rotatebox[origin=c]{90}{ETTh1}} &96 &\text{0.365}$\pm$2e-4 &\text{0.394}$\pm$2e-4&0.373 &0.398&0.37 &0.399&0.37 &0.4&0.375 &0.405&\underline{0.371} &\underline{0.394}&0.379 &0.409\\
         &192 &\text{0.398}$\pm$5e-4 &\text{0.417}$\pm$3e-4&0.405 &0.427&0.406 &0.428&0.413 &0.429&\underline{0.408} &\underline{0.423}&0.414 &0.423&0.411 &0.43\\
         & 336 &\text{0.421}$\pm$6e-4 &\text{0.422}$\pm$3e-4&0.427 &0.441&0.436 &0.451&\underline{0.422} &\underline{0.44}&0.435 &0.444&0.442 &0.445&0.43 &0.443\\
         & 720  &\text{0.417}$\pm$7e-3 &\text{0.439}$\pm$4e-3&\underline{0.442} &\underline{0.463}&0.472 &0.49&0.447 &0.468&0.457 &0.469&0.465 &0.472&0.446 &0.465\\

        \midrule[0.5pt]
        \multirow{4}{*}{\rotatebox[origin=c]{90}{ETTh2}} &96 &\text{0.269}$\pm$5e-4 &\text{0.336}$\pm$3e-4&0.278 &0.344&0.289 &0.353&\underline{0.274} &\underline{0.337}&0.286 &0.347&0.284 &0.348&0.275 &0.343\\
         &192 &\text{0.325}$\pm$9e-4 &\text{0.372}$\pm$6e-4&0.338 &0.382&0.383 &0.418&0.341 &0.382&0.347 &0.384&0.357 &0.4&0.337 &0.384\\
         & 336 &\text{0.348}$\pm$5e-4 &\text{0.394}$\pm$4e-4&\underline{0.356} &\underline{0.405}&0.448 &0.465&0.359 &0.405&0.374 &0.41&0.377 &0.417&0.364 &0.414\\
         & 720 &\text{0.381}$\pm$8e-4 &\text{0.426}$\pm$9e-4&0.391 &0.433&0.605 &0.551&\underline{0.388} &\underline{0.427}&0.406 &0.44&0.439 &0.456&0.397 &0.438\\

        \midrule[0.5pt]
        \multirow{4}{*}{\rotatebox[origin=c]{90}{ETTm1}} &96 &\text{0.291}$\pm$2e-3 &\text{0.346}$\pm$2e-3&\underline{0.293} &\underline{0.343}&0.305 &0.353&0.297 &0.348&0.295 &0.35&0.302 &0.349&0.293 &0.347\\
         &192 &\text{0.329}$\pm$8e-4 &\text{0.364}$\pm$2e-3 &\underline{0.331} &\underline{0.365}&0.33 &0.369&0.333 &0.376&0.332 &0.369&0.338 &0.373&0.333 &0.371\\
         & 336 &\text{0.354}$\pm$1e-3 &\text{0.384}$\pm$9e-4&0.363 &0.384&\underline{0.36} &\underline{0.384}&0.359 &0.392&0.365 &0.391&0.365 &0.385&0.364 &0.391\\
         & 720 &\text{0.393}$\pm$6e-4 &\text{0.407}$\pm$3e-4&0.405 &0.419&0.405 &0.413&\underline{0.397} &\underline{0.42}&0.416 &0.424&0.42 &0.42&0.42 &0.425\\

      \midrule[0.5pt]
        
        \multirow{4}{*}{\rotatebox[origin=c]{90}{ETTm2}} &96&\text{0.161}$\pm$2e-4 &\text{0.254}$\pm$5e-4&0.166 &0.258&0.163 &0.259&\underline{0.163} &\underline{0.255}&0.169 &0.261&0.165 &0.256&0.172 &0.255\\
         &192 &\text{0.213}$\pm$2e-3 &\text{0.295}$\pm$2e-3&0.222 &0.296&0.218 &0.302&\underline{0.216} &\underline{0.296}&0.227 &0.3&0.222 &0.296&0.231 &0.298\\
         & 336 &\text{0.259}$\pm$1e-3 &\text{0.326}$\pm$1e-3&0.274 &0.328&0.27 &0.34&\underline{0.266} &\underline{0.329}&0.274 &0.329&0.277 &0.333&0.276 &0.328\\
         & 720 &\text{0.330}$\pm$3e-3 &\text{0.373}$\pm$6e-4&0.34 &0.38&0.368 &0.406&\underline{0.339} &\underline{0.379}&0.352 &0.384&0.371 &0.389&0.349 &0.383\\

     \midrule[0.5pt]
        \multirow{4}{*}{\rotatebox[origin=c]{90}{Weather}} &96 &\text{0.145}$\pm$3e-3 &\text{0.194}$\pm$2e-3&\underline{0.146} &\underline{0.197}&0.165 &0.224&0.149 &0.198&0.146 &0.198&0.199 &0.262&0.152 &0.208\\
         &192 &\text{0.188}$\pm$4e-3 &\text{0.236}$\pm$4e-3&0.192 &0.24&0.207 &0.263&0.194 &0.241&\underline{0.19} &\underline{0.24}&0.228 &0.288&0.197 &0.251\\
         & 336 &\text{0.235}$\pm$2e-3 &\text{0.278}$\pm$8e-4&\underline{0.243} &\underline{0.279}&0.249 &0.294&0.244 &0.282&0.242 &0.283&0.267 &0.323&0.253 &0.296\\
         & 720 &\text{0.296}$\pm$1e-3 &\text{0.326}$\pm$9e-4&0.316 &0.332&0.308 &0.344&\underline{0.307} &\underline{0.33}&0.314 &0.333&0.319 &0.361&0.311 &0.343\\
  
       \midrule[0.5pt]
    
        \multirow{4}{*}{\rotatebox[origin=c]{90}{Electricity}} &96 &\text{0.127}$\pm$3e-4 &\text{0.222}$\pm$2e-4&\underline{0.131} &\underline{0.227}&0.133 &0.232&0.135 &0.231&0.133 &0.229&0.154 &0.267&0.143 &0.247\\
         &192 &\text{0.143}$\pm$8e-5 &\text{0.238}$\pm$6e-5&\underline{0.147} &\underline{0.242}&0.15 &0.249&0.15 &0.244&0.15 &0.245&0.164 &0.258&0.161 &0.266\\
         & 336 &\text{0.157}$\pm$6e-4 &\text{0.254}$\pm$2e-4&\underline{0.163} &\underline{0.259}&0.165 &0.267&0.166 &0.261&0.168 &0.264&0.188 &0.283&0.179 &0.285\\
         & 720 &\text{0.188}$\pm$3e-4 &\text{0.283}$\pm$5e-4 &0.201 &0.292&0.2 &0.302&0.206 &0.294&0.205 &0.296&0.236 &0.332&0.214 &0.318\\

      \midrule[0.5pt]
        \multirow{4}{*}{\rotatebox[origin=c]{90}{ILl}} &24 &\text{2.411}$\pm$0.012 &\text{1.085}$\pm$0.005&\underline{2.516} &\underline{1.116}&3.384 &1.359&2.975 &1.249&2.532 &1.103&3.699 &1.338&2.708 &1.147\\
         &36 &\text{2.324}$\pm$0.057 &\text{1.06}$\pm$0.015&2.569 &1.12&3.264 &1.311&2.91 &1.234&2.709 &1.125&3.128 &1.255&\underline{2.392} &\underline{1.066}\\
         & 48 &\text{2.352}$\pm$0.062 &\text{1.084}$\pm$0.003&2.746 &1.156&3.355 &1.334&2.925 &1.227&\underline{2.304} &\underline{1.048}&3.274 &1.268&2.39 &1.063\\
         & 60 &\text{2.453}$\pm$0.064 &\text{1.091}$\pm$0.015&2.778 &1.312&4.008 &1.464&3.028 &1.24&\underline{2.507} &\underline{1.079}&3.378 &1.303&2.621 &1.072\\
        
         \midrule[0.5pt]
        \multirow{4}{*}{\rotatebox[origin=c]{90}{Exchange}} &96 &\text{0.086}$\pm$8e-4 &\text{0.206}$\pm$0.001&0.095 &0.216&\underline{0.082} &\underline{0.205}&0.095 &0.219&0.097 &0.226&0.116 &0.246&0.1 &0.228\\
         &192 &\text{0.176}$\pm$0.001 &\text{0.301}$\pm$0.002&0.208 &0.324&\underline{0.167} &\underline{0.305}&0.189 &0.314&0.21 &0.332&0.228 &0.342&0.196 &0.322\\
         & 336 &\text{0.322}$\pm$0.007 &\text{0.408}$\pm$0.009&0.386 &0.447&0.348 &0.447&\underline{0.355} &\underline{0.435}&0.446 &0.488&0.398 &0.459&0.421 &0.482\\
         & 720 &\text{0.793}$\pm$0.04 &\text{0.674}$\pm$0.02&1.036 &0.766&0.894 &0.705&0.9 &0.708&1.001 &0.753&0.98 &0.75&0.962 &0.739\\
        \midrule[0.5pt]
        \multirow{4}{*}{\rotatebox[origin=c]{90}{Solar Energy}} &96 &\text{0.167}$\pm$0.003 &\text{0.229}$\pm$9e-4&0.174 &0.231&0.206 &0.271&\underline{0.166} &\underline{0.228}&0.187 &0.248&0.233 &0.259&0.176 &0.231\\
         &192 &\text{0.179}$\pm$5e-4 &\text{0.242}$\pm$0.002&0.185 &0.244&0.226 &0.293&\underline{0.179} &\underline{0.245}&0.216 &0.289&0.275 &0.284&0.195 &0.252\\
         & 336 &\text{0.188}$\pm$0.001 &\text{0.249}$\pm$0.001&\underline{0.19} &\underline{0.251}&0.241 &0.302&0.191 &0.254&0.213 &0.273&0.323 &0.309&0.2 &0.255\\
         & 720 &\text{0.2}$\pm$3e-4 &\underline{0.259}$\pm$0.001&\underline{0.2} &\text{0.258}&0.249 &0.309&0.203 &0.26&0.239 &0.291&0.335 &0.315&0.204 &0.262\\
         \midrule[0.5pt]
        \multirow{4}{*}{\rotatebox[origin=c]{90}{Traffic}} &96  &\underline{0.361}$\pm$3e-4 &\text{0.256}$\pm$4e-4&\text{0.357} &\underline{0.259}&0.4 &0.287&{0.373} &{0.267}&0.361 &0.265&0.416 &0.294&0.382 &0.261\\
         &192  &\underline{0.379}$\pm$8e-4 &\text{0.263}$\pm$6e-4&\text{0.376} &{0.270}&0.411 &0.291&0.384 &\underline{0.269}&{0.379} &{0.270}&0.408 &0.288&0.393 &0.275\\
         & 336 &0.39$\pm$6e-4 &\text{0.269}$\pm$7e-4&\underline{0.388} &{0.274}&0.425 &0.298&0.399 &0.275&\text{0.386} &\underline{0.271}&0.425 &0.298&0.433 &0.321\\
         & 720 &\text{0.421}$\pm$3e-5 &\text{0.283}$\pm$5e-4&\underline{0.423} &\underline{0.289}&0.465 &0.322&0.439 &0.295&0.432 &0.295&0.52 &0.353&0.474 &0.341\\

        \hline

    \end{tabular}}
% \vspace{-0.4cm}
\end{table*}

 \begin{table*}[h]
 % \vspace{-0.2cm}
    % \renewcommand{\arraystretch}{0.5}
    \setlength{\tabcolsep}{2.5pt}
    % {|>{\setlength{\tabcolsep}{3pt}}c|c|c|}
    \centering
    \caption{Full results of forecasting longer horizons 1020, 1320, and 1620 with input lengths 2048. ``-'' denotes out of memory. }
    % \vspace{-0.2cm}
    \label{tab:ultra_long_appendix_2048}
    % \begin{tabular}{c|c|p{20pt}p{20pt}|cc|cc|cc|cc|cc}
    {\footnotesize
    \begin{tabular}{c|c|cc|cc|cc|cc|cc|cc|cc|cc|cc|cc|cc}
        \hline
        \multirow{2}{*}{\shortstack{}} & &  \multicolumn{2}{c|}{SEMixer} &  \multicolumn{2}{c|}{DeformableTST} & \multicolumn{2}{c|}{TimeXer} & \multicolumn{2}{c|}{ModernTCN}& \multicolumn{2}{c|}{iTransformer}
        &  \multicolumn{2}{c|}{TimesNet}
        & \multicolumn{2}{c|}{TSMixer}& \multicolumn{2}{c|}{DLinear}& \multicolumn{2}{c|}{PatchTST}& \multicolumn{2}{c|}{TimeMixer}& \multicolumn{2}{c}{Scaleformer}\\
         & & MSE & MAE & MSE & MAE & MSE & MAE  & MSE & MAE & MSE & MAE& MSE & MAE& MSE & MAE& MSE & MAE& MSE & MAE& MSE & MAE& MSE & MAE\\ 
         \midrule[0.5pt]
\multirow{3}{*}{\rotatebox[origin=c]{90}{ETTh1}}&1020&\text{0.514} &\text{0.509}&0.573 &0.538&0.627 &0.575&0.695 &0.594&0.706 &0.632&1.164 &0.827&\underline{0.514} &\underline{0.509}&0.586 &0.565&0.517 &0.512&0.639 &0.575&0.648 &0.571\\
&1320&\underline{0.587} &\underline{0.546}&0.671 &0.587&0.722 &0.626&0.83 &0.649&0.833 &0.698&1.27 &0.883&\text{0.586} &\text{0.546}&0.663 &0.608&0.59 &0.548&0.858 &0.667&0.754 &0.617\\
&1620&\text{0.653} &\text{0.583}&0.745 &0.624&0.822 &0.686&1.005 &0.733&0.969 &0.761&1.439 &0.948&0.675 &0.585&0.73 &0.643&\underline{0.655} &\underline{0.583}&1.03 &0.747&0.955 &0.743\\
\midrule[0.5pt]
\multirow{3}{*}{\rotatebox[origin=c]{90}{ETTh2}}&1020&\text{0.452} &\text{0.483}&0.524 &0.534&0.47 &0.49&0.502 &0.507&0.49 &0.51&0.58 &0.558&\underline{0.455} &\underline{0.492}&1.136 &0.731&0.47 &0.502&0.523 &0.537&0.496 &0.523\\
&1320&\text{0.499} &\text{0.511}&0.537 &0.538&\underline{0.512} &\underline{0.513}&0.598 &0.549&0.538 &0.532&0.638 &0.596&0.508 &0.519&1.279 &0.779&0.515 &0.527&0.694 &0.622&0.575 &0.568\\
&1620&0.547 &0.538&0.579 &0.561&\text{0.506} &\text{0.501}&0.757 &0.628&0.576 &0.552&0.665 &0.588&\underline{0.53} &\underline{0.534}&1.237 &0.758&0.539 &0.538&0.614 &0.582&0.623 &0.592\\
\midrule[0.5pt]
\multirow{3}{*}{\rotatebox[origin=c]{90}{ETTm1}}&1020&0.418 &0.427&0.432 &0.435&0.457 &0.46&0.443 &0.453&0.458 &0.464&- &-&\underline{0.415} &\underline{0.427}&\text{0.413} &\text{0.423}&0.436 &0.442&0.464 &0.464&0.429 &0.441\\
&1320&0.439 &0.442&0.441 &0.441&0.461 &0.466&0.463 &0.465&0.479 &0.483&- &-&\text{0.426} &\text{0.435}&\underline{0.428} &\underline{0.433}&0.442 &0.448&0.496 &0.487&0.433 &0.442\\
&1620&0.457 &0.455&\underline{0.439} &\underline{0.442}&0.459 &0.469&0.472 &0.473&0.494 &0.489&- &-&\text{0.43} &\text{0.439}&0.438 &0.443&0.443 &0.45&0.504 &0.491&0.434 &0.447\\
\midrule[0.5pt]
\multirow{3}{*}{\rotatebox[origin=c]{90}{ETTm2}}&1020&\text{0.349} &\text{0.39}&0.382 &0.409&0.376 &0.409&0.385 &0.415&0.395 &0.419&- &-&\underline{0.364} &\underline{0.399}&0.424 &0.446&0.367 &0.402&0.412 &0.427&0.39 &0.428\\
&1320&\text{0.359} &\text{0.399}&0.38 &0.414&0.378 &0.416&0.39 &0.42&0.406 &0.43&- &-&\underline{0.371} &\underline{0.409}&0.467 &0.469&0.373 &0.41&0.416 &0.432&0.393 &0.434\\
&1620&\text{0.354} &\text{0.399}&0.374 &0.415&0.372 &0.414&0.39 &0.429&0.405 &0.432&- &-&\underline{0.368} &\underline{0.411}&0.501 &0.495&0.368 &0.411&0.413 &0.432&0.386 &0.431\\
\midrule[0.5pt]
\multirow{3}{*}{\rotatebox[origin=c]{90}{Weather}}&1020&\text{0.316} &\text{0.344}&\underline{0.327} &\underline{0.349}&0.332 &0.357&0.354 &0.378&0.356 &0.369&- &-&0.337 &0.364&0.321 &0.358&0.326 &0.354&0.366 &0.381&0.334 &0.367\\
&1320&\text{0.329} &\text{0.353}&\underline{0.336} &\underline{0.356}&0.342 &0.365&0.362 &0.384&0.347 &0.367&- &-&0.346 &0.37&0.333 &0.367&0.336 &0.361&0.374 &0.391&0.343 &0.374\\
&1620&\text{0.337} &\text{0.36}&0.346 &0.367&0.348 &0.37&0.368 &0.39&0.35 &0.372&- &-&0.353 &0.377&0.341 &0.376&\underline{0.344} &\underline{0.368}&0.405 &0.416&0.352 &0.382\\
\hline

    \end{tabular}}
% \vspace{-0.4cm}
\end{table*}

\end{document}